\documentclass[accepted]{uai2021} 

\usepackage[american]{babel}
\usepackage{microtype}
\usepackage{graphicx}
\usepackage{subcaption}
\usepackage[table,xcdraw]{xcolor}
\usepackage{booktabs} 
\usepackage{multirow}
\usepackage{bbm}
\usepackage{hyperref}
\usepackage[cmex10]{amsmath}
\usepackage{amssymb}
\usepackage{mathtools}
\usepackage[ruled,vlined]{algorithm2e}
\usepackage{algpseudocode}%
\usepackage{natbib} 
\usepackage{mathtools} 

\usepackage{booktabs} 
\usepackage{tikz} 

\newcommand{\xhdr}[1]{\vspace{0em}\noindent{{\bf #1.}}}
\newcommand{\ie}{\textit{i.e., \xspace}}
\newcommand{\eg}{\textit{e.g., \xspace}}
\newcommand{\hide}[1]{}

\newcommand{\std}[1]{\scriptsize{$\pm$#1}}
\newcommand{\indep}{\perp \!\!\! \perp}
\newcommand{\nifty}{\textsc{Nifty}\xspace}
\newcommand{\enc}{\textsc{Enc}\xspace}
\usepackage{colortbl}

\definecolor{Gray}{gray}{0.9}
\definecolor{LightCyan}{rgb}{0.88,1,1}
\newcolumntype{a}{>{\columncolor{Gray}}c}
\newcolumntype{b}{>{\columncolor{white}}c}

\graphicspath{{./FIG/}}

\title{Towards a Unified Framework for Fair and Stable\\ Graph Representation Learning}
\author[1]{\href{mailto:Chirag Agarwal <chiragagarwall12@gmail.com>?Subject=Your UAI 2021 paper}{Chirag~Agarwal}{}} 
\author[1]{Himabindu Lakkaraju\thanks{Equal Contribution}}
\author[1]{Marinka Zitnik\footnote[1]}
\affil[1]{%
    Harvard University
}

\begin{document}
\maketitle

\begin{abstract}
As the representations output by Graph Neural Networks (GNNs) are increasingly employed in real-world applications, it becomes important to ensure that these representations are fair and stable. 
In this work, we establish a key connection between counterfactual fairness and stability and leverage it to propose a novel framework, \nifty (uNIfying Fairness and stabiliTY), which can be used with any GNN to learn fair and stable representations. We introduce a novel objective function that simultaneously accounts for fairness and stability and develop a layer-wise weight normalization using the Lipschitz constant to enhance neural message passing in GNNs. In doing so, we enforce fairness and stability both in the objective function as well as in the GNN architecture. 
Further, we show theoretically that our layer-wise weight normalization promotes counterfactual fairness and stability in the resulting representations. We introduce three new graph datasets comprising of high-stakes decisions in criminal justice and financial lending domains. Extensive experimentation with the above datasets demonstrates the efficacy of our framework.
\end{abstract}
\section{Introduction}
\label{sec:intro}
Over the past decade, there has been a surge of interest in leveraging GNNs for graph representation learning. 
GNNs have been used to learn powerful representations that enabled critical predictions in downstream applications---\eg predicting protein-protein interactions~\citep{gainza2020deciphering,huang2020skipgnn}, drug repurposing~\citep{gysi2020network,zitnik2018modeling}, crime forecasting~\citep{jin2020addressing}, news and product recommendations~\citep{ying2018graph}.
As GNNs are increasingly implemented in real-world applications, it becomes important to ensure that these models and the resulting representations are safe and reliable. More specifically, it is important to ensure that these models and the representations they produce are not perpetrating undesirable discriminatory biases (\ie they are fair), and are also robust to attacks resulting from small perturbations to the graph structure and node attributes (\ie they are stable). 
\begin{figure}[t]
    \centering
    \includegraphics[width=0.49\textwidth]{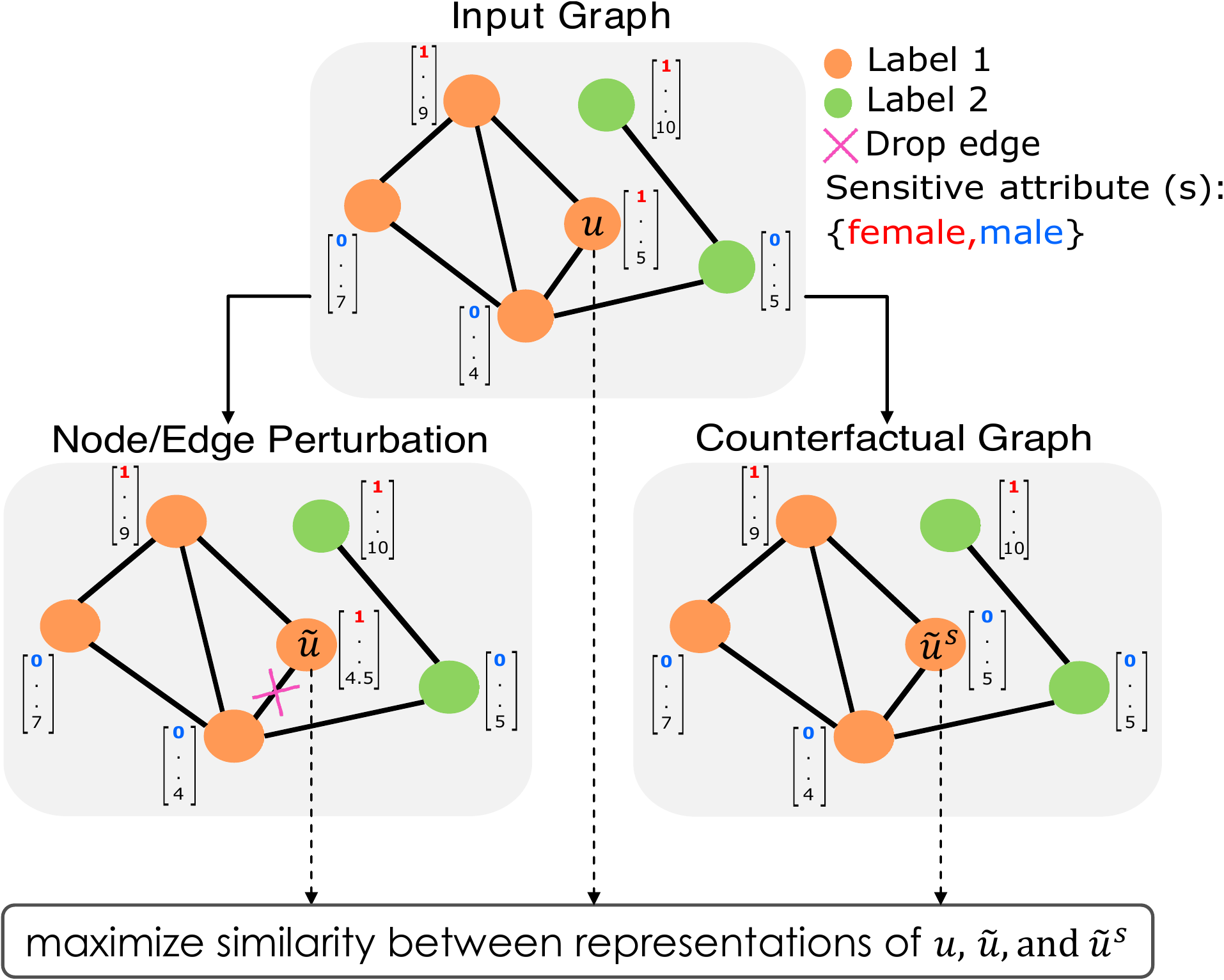}
    \caption{
    Our framework \nifty can learn node representations that are both fair and stable (\ie invariant to the sensitive attribute value and perturbations to the graph structure and non-sensitive attributes) by maximizing the similarity between representations from diverse augmented graphs.
    }
    \label{fig:nifty}
\end{figure}

A myriad of GNN methods with various neighborhood aggregation schemes have recently been developed~(\eg \cite{kipf17:semi,hamilton2017inductive,xu2018representation,xu2018powerful,velivckovic2018deep}).
While these methods achieve state-of-the-art performance in tasks such as node classification and link prediction, these methods can be prone to discrimination and instability~\citep{dai2020fairgnn,rahman2019fairwalk,bose2019compositional}. Furthermore, prior work has argued that GNNs not only capture the undesirable biases prevalent in the data, but may also exacerbate them thanks to their message passing schemes~\citep{dai2020fairgnn}. 
Generally, in graphs such as social networks, nodes with similar sensitive attribute (\eg race, age) values are likely to connect to each other~\citep{dai2020fairgnn}. 
Since GNNs compute node representations by propagating and aggregating neural messages along edges in graph neighborhoods, nodes with similar sensitive attribute values are likely to share similar representations leading to severe discriminatory biases, \ie downstream predictions may be highly correlated with sensitive attributes. 

Recent research has treated fairness and stability in GNNs as independent problems and proposed standalone solutions for the same. For example,~\citet{dai2020fairgnn} proposed FairGNN to promote fairness in GNNs through an objective function that incorporates group fairness measures such as statistical parity and equality of opportunity. On the other hand,~\citet{zhu2019robust} aimed to make GNNs stable and robust to adversarial attacks. While these techniques provide a promising approach to study fairness and stability independently, it remains an open question whether there are any deeper connections between fairness and stability in GNNs, and if these properties can be achieved simultaneously.

\xhdr{Present work}
Here, we address the problem of learning node representations that are both fair and stable. To tackle this problem, we first identify a key connection between counterfactual fairness and stability. While stability accounts for robustness w.r.t. small random perturbations to node attributes and/or edges, counterfactual fairness accounts for robustness w.r.t. modifications of the sensitive attribute. We leverage this connection to propose a novel framework, \nifty (uNIfying Fairness and stabiliTY), that can be used with any existing GNN model to learn fair and stable representations. Our framework exploits the aforementioned connection  
to enforce fairness and stability both in the objective function as well as in the GNN architecture. More specifically,  
we introduce a novel objective function which simultaneously optimizes for counterfactual fairness and stability 
by maximizing the similarity between representations of the original nodes in the graph, and their counterparts in the augmented graph (Fig.~\ref{fig:nifty}). Nodes in the augmented graph are generated by slightly perturbing the original node attributes and edges or by considering counterfactuals of the original nodes where the value of the sensitive attribute is modified. We also develop a novel method for improving neural message passing
by carrying out layer-wise weight normalization using the Lipschitz constant. We theoretically show that this normalization promotes counterfactual fairness and stability of learned representations. To the best of our knowledge, this work is the first to tackle the problem of learning node representations that are both fair and stable. 

We introduce and experiment with three new graph datasets comprising of critical decisions in criminal justice (if a defendant should be released on bail) and financial lending (if an individual should be given loan) domains. Our results show that \nifty improves the fairness and stability of five GNNs
by 92.01\% and 60.87\% respectively (on an average) without sacrificing predictive performance. We also observe that the resulting representations become fairer not only w.r.t. the notion of counterfactual fairness but also w.r.t. other notions of group fairness such as statistical parity and equality of opportunity. Further, our results establish that enforcing fairness and stability both in the objective function as well as in the GNN architecture can be incredibly beneficial for learning fair and stable representations. 

\section{Related Work}
\label{sec:related}
This work lies at the intersection of fairness and stability in machine learning, and Graph Neural Networks (GNNs). Below we discuss related work for each of these topics. 

\xhdr{Fairness} 
Several competing and contrasting notions of fairness have been proposed in recent literature. They can be broadly categorized into: 1) \emph{group fairness}, which emphasizes that minority groups should receive similar treatment as that of advantaged groups~\citep{BerkHJKR18,hardt16}, 2) \emph{individual fairness}, which requires that \emph{similar} individuals should be treated similarly~\citep{dwork12}, and 3) \emph{counterfactual fairness}, which captures the intuition that a decision pertaining to an individual is fair if 
changing the individual's sensitive attribute value does not affect the decision~\citep{kusner2017counterfactual}. Furthermore, various metrics have been proposed to realize each of the aforementioned notions of fairness. For example, statistical (demographic) parity, equalized odds, equality of opportunity, and predictive parity are metrics proposed to enforce group fairness. These metrics have also been leveraged to develop new objective functions for constructing machine learning models that are both fair and accurate~\citep{zafar2017parity,zafar2017fairness}.
Prior research has also established that certain notions of fairness (calibration and balance conditions) are fundamentally incompatible and cannot be simultaneously optimized~\citep{KleinbergMR17,Chouldechova17}. 

\xhdr{Graph Neural Networks} 
Deep learning on graphs and GNNs, in particular, learn how to represent nodes in a graph as points,  \ie embeddings, in a vector embedding space, where the geometry of the embedding space is optimized to reflect topology of the graph as well as node attribute information~\citep{wu2020comprehensive}. 
Motivated by spectral graph convolutions \citep{hammond2011wavelets,defferrard2016convolutional}, Graph Convolutional Networks (GCN)~\citep{kipf17:semi} specified deep transformation functions akin to applying convolutional filters over local graph neighborhoods.
The subsequent methods, \eg \citet{gilmer2017neural,hamilton2017inductive,hu2020heterogeneous,lee2019self,alsentzer2020subgraph}, developed efficient algorithms for rich types of graphs and larger structures, including edges, subgraphs, and entire graphs by generating embeddings through a series of transformations that exchange embeddings between neighboring nodes in the graph.
For example, Jumping Knowledge (JK) Networks~\citep{xu2018representation} use skip connections to leverage diverse local neighborhoods and generate richer representations.
Similarly, Graph Isomorphism Networks (GIN)~\citep{xu2018powerful} adaptively adjust the importance weights of nodes and Deep Graph Infomax (DGI)~\citep{velivckovic2018deep} relies on maximizing mutual information between patch representations and high-level graph summaries to produce node representations.

\xhdr{Fairness and Stability in GNNs}
Recent studies addressed the issues of fairness and stability in GNNs~\citep{dai2020fairgnn,fisher2020debiasing,geisler2020reliable,bose2019compositional,rahman2019fairwalk,zhu2019robust,zhang2020gnnguard}.
To achieve fairness, existing work de-biases embeddings with respect to sensitive attributes via adversarial learning frameworks~\citep{dai2020fairgnn,bose2019compositional}.
These methods use regularization to implement the notion of group fairness; however, they are incapable of achieving counterfactual fairness.
To achieve stability, recent methods use adversarial training~\citep{zugner2019adversarial}, robust message-aggregation~\citep{geisler2020reliable}, and attention mechanisms~\citep{zhu2019robust} to defend GNNs against a variety of attacks that perturb discrete graph structure or node attributes.
In contrast, our unifying framework can learn graph embeddings that are simultaneously fair and stable.

\section{Preliminaries}
\label{sec:prelims}
\xhdr{Notation} Let $\mathcal{G} = (\mathcal{V}, \mathcal{E}, \mathbf{X})$ denote an undirected graph comprising of a set of nodes $\mathcal{V}$ and a set of edges $\mathcal{E}$. Let $\mathbf{X} = \{\mathbf{x}_{1}, \mathbf{x}_{2}, \dots, \mathbf{x}_{N}\}$ denote the set of node attribute vectors corresponding to all the nodes in $\mathcal{V}$. More specifically, $\mathbf{x}_{v} \in \mathbf{X}$ is an $M$-dimensional vector which captures the attribute values of node $v \in \mathcal{V}$.
Let $N = |\mathcal{V}|$ denote the number of nodes in the graph and let $\mathbf{A} \in \mathbb{R}^{N \times N}$ be the graph adjacency
matrix where element $\mathbf{A}_{uv} = 1$ if there exists some edge $e \in \mathcal{E}$ between nodes $u$ and $v$, and $\mathbf{A}_{uv} = 0$ otherwise.
We also use $\mathcal{N}_{u}$ to denote the set of immediate neighbors of node $u$, \ie $\mathcal{N}_{u} = \{ v \in \mathcal{V} | A_{uv} = 1 \}$. Furthermore, let $\mathbf{I}_{u} \in \{0,1\}^{N}$ denote the binary incidence vector which captures all the edges incident on node $u$, \ie $\mathbf{I}_{uv} = 1$ if an edge exists between nodes $u$ and $v$ otherwise it is set to $0$. Finally, we introduce $\textbf{b}_u$ to capture all the information associated with node $u$, \ie $\textbf{b}_u = [ \textbf{x}_u; \textbf{I}_u]$ denotes the concatenation of node attribute vector and binary incidence vector corresponding to node $u$.  We also generate an augmented graph $\mathcal{G'} = (\mathcal{V}, \mathcal{E'}, \mathbf{\Tilde{X}})$ as follows: for each node $u \in \mathcal{V}$ in the original graph, we generate a corresponding node in the augmented graph by slightly perturbing the attribute values, incident edges, and/or modifying the value of the sensitive attribute of node $u$. The adjacency matrix and node attribute vectors corresponding to this augmented graph $\mathcal{G'}$ are denoted by  $\mathbf{\Tilde{A}}~\text{and}~\mathbf{\Tilde{X}}$. 

We consider a GNN with $K$ layers and denote the representations output by each of these layers as $\textbf{h}^{1}_{u}, \textbf{h}^{2}_{u}, \cdots \textbf{h}^{K-1}_{u}, \textbf{h}^{K}_{u} $ for a given node $u$. We use $\mathbf{z}_{u}$ to denote the representation output by the last layer of the GNN for node $u$ \ie $\mathbf{z}_{u} = \textbf{h}^{K}_{u} $.
Analogously, $\mathbf{\Tilde{z}}_{u}$ denotes the representation output by the last layer of the GNN for node $u$ in the augmented graph $\mathcal{G}'$. We assume that the (dis)similarity between any two node representations is given by a distance metric $D: \mathbb{R}^{d} \times \mathbb{R}^{d} \to \mathbb{R}$. 
Our goal is to learn an encoder function \enc which maps a given node $u$ to a representation $\mathbf{z}_{u}$ \ie $\enc(u) = \mathbf{z}_{u}$.
Lastly, let $f$ denote a downstream classifier that maps the node representation $\mathbf{z}_{u}$ of a given node $u$ to a class label $\hat{y}_{u}$. 
\hide{
In a node classification task, a classifier $f: \mathbf{z}_{u} \to \hat{y}_{u}$ maps the node representation $\mathbf{z}_{u}$ of node $u$ to its predicted class.
Finally, $\mathbf{\Tilde{z}}_{u}$ denote the representations of node $u$ in an augmented graph $\mathcal{G'} \in (\mathcal{V, \mathcal{E'}, \mathbf{\Tilde{X}})}$, where $\mathbf{\Tilde{A}}~\text{and}~\mathbf{\Tilde{X}}$ are the perturbed adjacency and node-attribute matrix.
We aim to learn a function $h:\mathbf{x}_{u}\to\mathbb{R}^{d}$ that maps node attributes to node embeddings $\mathbf{z}_{u} = h(\mathbf{x}_{u})$, where $\mathbf{z}_{u}$ is the embedding at the final layer $K$ of the encoder $h$.
In this case, the similarity between two embeddings can be given by a distance metric $D: \mathbb{R}^{d} \times \mathbb{R}^{d} \to \mathbb{R}$, \ie it takes two embeddings 
$\mathbf{z}_{u},\mathbf{z}_{v} \in \mathbb{R}^{d}$ and outputs the similarity between them.
Further, $\mathbf{Z} = \{\mathbf{z}_{1}, \mathbf{z}_{2}, \dots, \mathbf{z}_{N}\}$ denotes the set of representations.

}

\xhdr{Graph Neural Networks}
Many GNNs can be formulated as message passing networks~\citep{wu2020comprehensive} specified by trainable operators \textsc{Msg}, \textsc{Agg}, and \textsc{Upd}. In a $K$-layer GNN, the operators are recursively applied on $\mathcal{G}$, specifying how neural messages (\ie embeddings) are exchanged between nodes, aggregated, and transformed to arrive at final node representations in the last layer of transformations. Typically, a message between a pair of nodes $(u, v)$ in layer $k$ is defined as a function of hidden representations of nodes $\mathbf{h}_u^{k-1}$ and $\mathbf{h}_v^{k-1}$ from the previous layer: $\mathbf{m}_{uv}^k = \textsc{Msg}(\mathbf{h}_u^{k-1}, \mathbf{h}_v^{k-1}).$ 
In \textsc{Agg}, messages from $\mathcal{N}_{u}$ are aggregated as $\mathbf{m}_{u}^{k}=\textsc{Agg}(\mathbf{m}_{uv}^{k} | u\in\mathcal{N}_u)$. In \textsc{Upd}, the aggregated message $\mathbf{m}_{u}^{k}$ is combined with $\mathbf{h}_u^{k-1}$ to produce $u$'s representation for layer $k$  as $\mathbf{h}_{u}^{k}=\textsc{Upd}(\mathbf{m}_{u}^{k}, \mathbf{h}_{u}^{k-1})$. Final node representation $\mathbf{z}_u = \mathbf{h}_u^K$ is the output of the last layer.

\xhdr{Fairness and Stability} Our goal is to learn node representations that are fair and stable. More specifically, the notions of fairness and stability that we consider in this work are counterfactual fairness and Lipschitz continuity respectively. Below, we provide definitions of these notions and formalize them in the context of graph representation learning.

\emph{\textbf{Counterfactual Fairness}}: A function is considered to be counterfactually fair if its output is independent of the sensitive attribute, \ie changing the sensitive attribute value of any given instance should not affect the output of the function for that instance. 
In the context of graph representation learning, this notion can be interpreted as follows: node representations output by encoders should be independent of the sensitive attribute. 
\xhdr{Definition 1} \textit{An encoder function \enc 
satisfies counterfactual fairness if the following holds for any given node $u$:
}
\begin{equation}
    \label{eq:counterfactual}
    \enc(u) = \enc(\Tilde{u}^{s})
\end{equation}
where $\Tilde{u}^{s}$ is a node in the augmented graph which is generated by modifying/flipping the value of the sensitive attribute (s) of node $u$ while keeping everything else constant. 

\hide{
\xhdr{Definition 1} \textit{Node representation $\mathbf{z}_{u}$ corresponding to a node $u$ in graph $\mathcal{G}$ satisfies counterfactual fairness w.r.t. a sensitive attribute $s \in \{0, 1\}$ 
if $\mathbf{z}_{u}$ is independent of $s$, \ie
}
\begin{equation}
    \label{eq:counterfactual}
    P(\mathbf{z}_{u} | s=0) = P(\mathbf{z}_{u} | s=1)
\end{equation}
}
\emph{\textbf{Stability via Lipschitz Continuity}}:
A function is considered to be stable according to the notion of Lipschitz continuity if slightly perturbing any given instance does not drastically change the output of the function. 
In the context of graph representation learning, this notion can be interpreted as follows: small perturbations to node attributes and/or incident edges should not drastically change the resulting representations. 

\hide{
\xhdr{Definition 2}
\textit{A function $h:\mathbb{R}^{M} \times \mathbb{R}^{N} \to\mathbb{R}^{d}$ 
which takes as input the node attribute vector ($\mathbf{x}_u$) and the incident edge vector ($\mathbf{A}_u$) of any given node $u$ and outputs the corresponding node representation ($\mathbf{z}_{u}$) is stable according to the notion of Lipschitz continuity if:
\begin{equation}
    \label{eq:stability}
    ||h(\mathbf{\Tilde{x}}_{u},\Tilde{\textbf{A}}_u) - h(\mathbf{x}_{u},\textbf{A}_u)||_{p} \leq L||(\mathbf{\mathbf{\Tilde{x}}_{u},\Tilde{\textbf{A}}_u}) - (\mathbf{x}_{u},\textbf{A}_u) ||_{p},
\end{equation}
}
where $\mathbf{\Tilde{x}}_{u}$ and $\Tilde{\textbf{A}}_u$ denote perturbations to the node attribute vector $\textbf{x}_u$ and incident edge vector $\textbf{A}_u$ respectively, and $L$ is the Lipschitz constant.
}

\xhdr{Definition 2}
\textit{An encoder function \enc is stable according to the notion of Lipschitz continuity if:}
\begin{equation}
    \label{eq:stability}
    ||\enc(\Tilde{u}) - \enc(u)||_{p} \leq L||\mathbf{\Tilde{b}}_{u} - \mathbf{b}_{u}||_{p},
\end{equation}
where $\Tilde{u}$ is a node in the augmented graph generated by perturbing $u$'s attribute values and/or incident edges,  $\mathbf{b}_{u}$ and $\mathbf{\Tilde{b}}_{u}$ capture the attribute and incident edge information for nodes $u$ and $\Tilde{u}$ respectively, and $L$ is the Lipschitz constant.

\section{Our Framework \nifty}
\label{sec:methods}
Next, we describe our framework \nifty which aims to generate fair and stable graph embeddings. To achieve this goal, \nifty infuses fairness and stability in the objective function (Section~\ref{sec:objective}) as well as in the architecture (Section~\ref{sec:architecture}) of underlying GNN. 

\xhdr{Problem formulation~(Fair and Stable embeddings)} \textit{Given a graph $\mathcal{G} = (\mathcal{V},\mathcal{E},\mathbf{X})$, \nifty aims to generate $d$-dim. embeddings $\mathbf{z}_{u}\in\mathbb{R}^{d}$ that are counterfactually fair (Eq.~\ref{eq:counterfactual}) and stable to attribute and structural perturbations of $\mathcal{G}$ (Eq.~\ref{eq:stability}).
}

\hide{
\nifty is a unified framework that is easy to incorporate into any GNN.
In essence, it requires $h$ to generate graph embeddings invariant to sensitive attributes and noise.
Note that while this paper focuses on the fairness and stability properties in a node-classification task, \nifty entails
learning an embedding function $h:\mathbb{R}^{M} \times \mathbb{R}^{N} \to\mathbb{R}^{d}$ that maps node attributes and incident edges to a lower dimensional representation that can be extended without loss of generality to other downstream-tasks.
}


\subsection{Enforcing Fairness and Stability in the Objective function} \label{sec:objective}
%



To infuse fairness and stability in the objective function, we introduce a triplet-based objective that maximizes the agreement between the original graph and its counterfactual and noisy views.
To this end, we build off the Siamese networks to maximize this agreement, \ie the two augmented network neighborhoods and the augmented attribute vectors of the same node should result in the same embedding~\citep{chen2020simple,chen2020exploring}.
Next, we describe the graph augmentation procedure.


Generating augmented views of graph structure and attribute information is key for the Siamese learning approach.
We generate them using node-, sensitive attribute-, and edge-level perturbations.

\xhdr{a) Perturbing node attributes} 
We draw a random attribute masking vector $\mathbf{r} \in \{0, 1\}^M$ from a Bernoulli distribution, \ie $\mathbf{r} \sim \mathcal{B}(p_{n})$, where $p_{n}$ is the probability of independently perturbing each attribute (except for the sensitive attribute $s$) in $\mathbf{x}_u$.
The augmented attribute vector is then defined as $\mathbf{\Tilde{x}}_{u} = \mathbf{x}_{u} + \mathbf{r} \circ \delta$, where $\delta \in \mathbb{R}^{M}$ is sampled from a normal distribution.

\xhdr{b) Counterfactual perturbation of sensitive attribute} 
We modify the value of sensitive attribute $s$ in $\mathbf{x}_{u}$ to generate a counterfactual. More specifically, we consider the case where the sensitive attribute is a binary variable (i.e., $s \in \{0,1\}$) and we create a counterfactual node $\tilde{u}^s$ by flipping the value of $s$ from 0 to 1 or vice-versa.

\xhdr{c) Perturbing graph structure}
We draw a random binary mask from a Bernoulli distribution, \ie $\mathbf{R}_{e} \sim \mathcal{B}(1-p_{e})$, where $\mathbf{R}_{e} \in \{0, 1\}^{N \times N}$ and $p_{e}$ denotes the probability with which an edge is dropped from $\mathcal{G}$.
We construct the augmented adjacency matrix as $\mathbf{\Tilde{A}} = \mathbf{A} \circ \mathbf{R}_{e}$.

To learn embeddings that are invariant to the sensitive attribute and stable against perturbations of the graph structure and non-sensitive attributes, we train the GNN encoder \enc using the Siamese framework~\citep{bromley1994signature}.
The encoder generates representations $\mathbf{\Tilde{z}}_{u}$ of the augmented graph at every iteration.
By generating augmented graphs, \nifty can induce appropriate bias into the underlying GNN to learn embeddings that are invariant to the combination of counterfactual nodes as well as to random perturbations in the graph structure. 
A predictor $t: \mathbb{R}^{d}\to\mathbb{R}^{d}$ consisting of a fully-connected neural layer is then used to transform and match the representations with each other. Inspired by \citet{grill2020bootstrap}, we define a triplet-based objective function that optimizes the similarity between the original graph and its augmented (\ie counterfactual and noisy) representations:
\begin{equation}
    \label{eq:sim_loss}
    \mathcal{L}_{s} = \mathbb{E}_{u}\big[ \frac{1}{2}\big(D(t(\mathbf{z}_{u}), \textrm{sg}(\mathbf{\Tilde{z}}_{u})) +  D(t(\mathbf{\Tilde{z}}_{u}), \textrm{sg}(\mathbf{z}_{u}))\big)\big],
\end{equation}
where $t(\mathbf{z}_{u})$ and $t(\Tilde{\mathbf{z}}_{u})$ are the transformed representations of node $u$ and perturbed node $\Tilde{u}$ respectively,
$D$ is the cosine distance, and stopgrad ($\textrm{sg}$) prevents gradients from being backpropagated. 
The \textrm{stopgrad} signifies that the node representations $\mathbf{\Tilde{z}}_{u}$ are considered as constant when operating on $t(\mathbf{z}_{u})$ and vice-versa.

Finally, the overall objective function for \nifty is:
\begin{equation}
    \label{eq:objective}
    \min_{\theta_{\enc}, \theta_{t}, \theta_{f}} \mathop{\mathbb{E}_{u}}\big[ (1-\lambda)\mathcal{L}_{c}] + \lambda \mathcal{L}_{s},
\end{equation}
where $\{\theta_{\enc}, \theta_{t}, \theta_{f}\}$ denotes trainable parameters of \enc, predictor $t$, and classifier $f$, $\mathcal{L}_{c}$ is the binary cross entropy (BCE) loss, and the expectation is taken over training nodes in $\mathcal{G}$.
The regularization coefficient $\lambda$ controls the trade-off between downstream node classification loss $\mathcal{L}_c$ and the tripled-based objective $\mathcal{L}_s$. Algorithm~\ref{alg:nifty} summarizes the overall training procedure of \nifty.


\subsection{Enforcing Fairness and Stability in GNN architecture} \label{sec:architecture}
Next, we describe how \nifty infuses fairness and stability in the architecture of the underlying GNN. In particular, \nifty modifies the GNN's routing of neural messages. Recall (Sec.~\ref{sec:prelims}) that a typical GNN layer is given by: $\mathbf{h}_{u}^{k}=\textsc{Upd}(\textsc{Agg}(\textsc{Msg}(\mathbf{h}_{u}^{k-1}, \mathbf{h}_{v}^{k-1}) | v \in \mathcal{N}_{u}), \mathbf{h}_{u}^{k-1})$. As we will see in this section, \nifty modifies the $\textsc{Upd}$ step of each GNN layer.

Without loss of generality, we can consider $\textsc{Agg}$ operator to be a fully-connected layer and $\textsc{Upd}$ to be a non-linear activation function $\sigma$. 
Using these specific parametrizations, the message-passing step can be rewritten as:~$\mathbf{h}_{u}^{k} = \sigma\big(\mathbf{W}_{a}^{k}~\mathbf{h}_{u}^{k-1}{+} \mathbf{W}_{n}^{k}\sum_{{v\in\mathcal{N}(u)}}\mathbf{h}_{v}^{k-1}\big)$, where $\mathbf{W}_{n}^{k}$ is the weight matrix associated with the neighbors of node $u$ at layer $k$ and $\mathbf{W}_{a}^{k}$ is the self-attention weight matrix at layer $k$. 

Definition~2 tells us that as the local network neighborhood and the node attribute vector of node $u$ change from  $\mathbf{b}_{u}$ to $\mathbf{\Tilde{b}}_{u}$, the Lipschitz constant $L$ provides an upper bound on how much $u$'s node embedding can change. In fact, the Lipschitz constant $L$ represents the smallest value for which Eqn.~\ref{eq:stability} in Definition~2 holds true.
Leveraging this understanding, \nifty bounds the change in $u$'s embedding by appropriately normalizing the encoder's weight matrices. This is possible because of the slope-restricted structure of the nonlinear activation function in the \textsc{Upd} step (see proof in Sec.~\ref{sec:theory}).
Using our derivations in Sec.~\ref{sec:theory}, at each layer $k$, we calculate the Lipschitz constant $L$ of term $\mathbf{W}_{a}^{k}\mathbf{h}_{u}^{k-1}$ as the spectral norm of the weight matrix. We use $L$ to normalize $\mathbf{W}_{a}^{k}$ as: 
\begin{equation}
    \mathbf{\Tilde{W}}_{a}^{k} = \mathbf{W}_{a}^{k}/\sigma(\mathbf{W}_{a}^{k}). \label{eq:lipshitz-norm}
\end{equation}
We use this Lipschitz-normalized weight matrix $\mathbf{\Tilde{W}}_{a}^{k}$ to modify the \textsc{Upd} step as:~
$\mathbf{h}_{u}^{k} = \sigma(\mathbf{\Tilde{W}}_{a}^{k}~\mathbf{h}_{u}^{k-1} + \mathbf{W}_{n}^{k}\sum_{{v\in\mathcal{N}(u)}}\mathbf{h}_{v}^{k-1})$.

Lipschitz normalization of weight matrices is appealing for two reasons. It bounds the difference between embeddings of original and perturbed nodes (attributes). It also establishes a connection between the stability and counterfactual fairness in a sense that similar inputs should yield similar predictions. Next, we investigate this connection in detail. 

\begin{algorithm}
\SetAlgoLined
\footnotesize
\DontPrintSemicolon
\textbf{Input}: Graph $\mathcal{G}=(\mathcal{V}, \mathcal{E}, \mathbf{X})$; regularization $\lambda$;
sensitive attribute $s$; number of training epochs $\textit{num\_epoch}$\\
\textbf{Output}: Optimized model parameters \{$\theta_{\enc}$, $\theta_{t}$, $\theta_{f}$\}; fair and stable representations $\mathbf{z}_{u}$ for $u \in \mathcal{G}$\\ 
 \For{$ep\gets1$ \KwTo \textit{num\_epoch}}{
    \For{$layer~k\gets1$ \KwTo \textit{K}}{
        Lipschitz-normalize \enc's weights $\mathbf{W}_{a}^{k}$ (Eqn.~\ref{eq:lipshitz-norm})
    }
    \For{$node~u\gets1$ \KwTo \textit{|$\mathcal{V}$|}}{
        Perturb attributes and graph structure to get $\Tilde{u}$ (Sec.~\ref{sec:objective})\!\!\!\! \\
        Modify sensitive attribute value to get $\Tilde{u}^{s}$
        (Sec.~\ref{sec:objective}) \\ 
        Encode $\mathbf{z}_{u}=\enc(u)$, $\mathbf{\Tilde{z}}_{u}=\enc(\Tilde{u})$, $\mathbf{\Tilde{z}}_{u}^{s}=\enc(\Tilde{u}^{s})$\\
        Transform embeddings: $t(\mathbf{z}_{u}),~t(\mathbf{\Tilde{z}}_{u}),~t(\mathbf{z}_{\Tilde{u}}^{s})$ (Sec.~\ref{sec:objective})\\
    }
    Calculate triplet-based similarity (Eqn.~\ref{eq:sim_loss}) \\
    Apply downstream classifier $f$ as $\hat{y}_{u}=f(\enc(u))$ \\
    Update $\{\theta_{\enc},\theta_{t},\theta_{f}\}$ according to the objective in Eqn.~\ref{eq:objective}\!\!\! \\
}
\caption{Overview of \nifty algorithm}
\label{alg:nifty}
\end{algorithm}

\section{Theoretical analysis of \nifty}
\label{sec:theory}
Here, we provide detailed theoretical analysis of our framework \nifty. More specifically, we prove that representations generated by \nifty are stable. We also provide a theoretical upper bound on the unfairness of the resulting representations. Lastly, we show that the downstream classifiers that leverage the representations output by \nifty satisfy counterfactual fairness as well. 

\xhdr{Theorem 1 (\nifty Stability)}~\textit{Given a non-linear activation function $\sigma$ that is Lipschitz continuous, the representations learned by our framework \nifty are stable i.e., 
} 
\begin{equation}\label{eqn:theorem1}
||\enc(\Tilde{u}) - \enc(u)||_{p} \leq \prod_{k=1}^{K}||\mathbf{W}_{a}^{k}||_{p}||(\Tilde{\mathbf{b}}_{u}-\mathbf{b}_{u})||_{p},
\end{equation}
where $\Tilde{u}$ is a node in the augmented graph which is generated by
perturbing the attribute values and/or incident edges of node $u$, 
$\mathbf{b}_{u}$ and $\mathbf{\Tilde{b}}_{u}$ capture all attribute values and incident edge information for nodes $u$ and $\Tilde{u}$ respectively, and $\mathbf{W}_{a}^{k}$ is weight matrix associated with attributes of node $u$ at layer $k$. 

\textit{Proof.~}
Following Sec.~\ref{sec:architecture}, the node representation output by layer $k$ of the GNN for a perturbed node $\Tilde{u}$ is given by:
\begin{equation}
    \label{eq:node_p}
    \Tilde{\mathbf{h}}_{u}^{k} = \sigma\big(\mathbf{W}_{a}^{k}~\Tilde{\mathbf{h}}_{u}^{k-1}\!+\!\mathbf{W}_{n}^{k}\sum_{\mathclap{v\in\mathcal{N}(\Tilde{u})}}\mathbf{h}_{v}^{k-1}\big),
\end{equation}
where $\mathcal{N}(\Tilde{u})$ is the neighborhood of node $\Tilde{u}$ which is obtained after perturbing edges incident on node $u$.
Now, the difference between the node embeddings obtained after the message-passing in layer $k$ is:
\vspace*{-\abovedisplayskip}
\begin{equation}
    \begin{aligned}
    & \mathbf{\Tilde{h}}_{u}^{k}{-}\mathbf{h}_{u}^{k} = 
    \\  
    & \sigma\big(\mathbf{W}_{a}^{k}\Tilde{\mathbf{h}}_{u}^{k{-}1}{+}\mathbf{W}_{n}^{k}\sum_{\mathclap{v\in\mathcal{N}(\Tilde{u})}}\mathbf{h}_{v}^{k-1}\big){-}\sigma\big(\mathbf{W}_{a}^{k}{\mathbf{h}}_{u}^{k{-}1}{+}\mathbf{W}_{n}^{k}\sum_{\mathclap{v\in\mathcal{N}(u)}}\mathbf{h}_{v}^{k{-}1}\big) \nonumber
    \end{aligned}
\end{equation}
Taking the norm and assuming that $\sigma$ is normalized Lipschitz, \ie $||\sigma{(b)}-\sigma{(a)}||_{p} \leq ||b-a||_{p}$, we get:
\vspace*{-\abovedisplayskip}
\begin{eqnarray}
    \begin{aligned}
    & ||\mathbf{\Tilde{h}}_{u}^{k}{-}\mathbf{h}_{u}^{k}||_{p} \\ 
    & \leq ||\textbf{W}_{a}^{k}(\mathbf{\Tilde{h}}_{u}^{k{-}1}{-}\textbf{h}_{u}^{k{-}1}){+}\textbf{W}_{n}^{k}(\sum_{\mathclap{v\in\mathcal{N}(\Tilde{u})}}\textbf{h}_{v}^{k{-}1}{-}\sum_{\mathclap{v\in\mathcal{N}(u)}}\textbf{h}_{v}^{k{-}1})||_{p}\\
    \end{aligned}
\end{eqnarray}
The second term in the above inequality will be close to 0 since the probability of dropping an edge $p_{e}$ is very small. So, we can drop the second term and then leverage Cauchy-Schwartz inequality to get: 
\begin{eqnarray}
    \begin{aligned}
    & \!\!\! ||\mathbf{\Tilde{h}}_{u}^{k}-\mathbf{h}_{u}^{k}||_{p} \leq ||\textbf{W}_{a}^{k}(\mathbf{\Tilde{h}}_{u}^{k-1}{-}\textbf{h}_{u}^{k{-}1})\\
    & \leq ||\mathbf{W}_{a}^{k}||_{p}||(\mathbf{\Tilde{b}}_{u}-\mathbf{b}_{u})||_{p} \!\!\!\\ 
    \end{aligned}
\end{eqnarray}
Note that the encoder \enc is essentially a sequential composition of message-passing functions applied at layers $1 \cdots K$. Furthermore, the composition of two Lipschitz continuous functions with Lipschitz constants $L_1$ and $L_2$ is a new Lipschitz continuous function with $L_1 \times L_2$ as the Lipschitz constant~\citep{gouk2021regularisation}. Putting it all together, we have:
\begin{eqnarray}
\begin{aligned}
   ||\enc(\Tilde{u}){-}\enc(u)||_{p} = 
   || \mathbf{\Tilde{z}}_u{-}\mathbf{z}_u||_{p} = 
   || \mathbf{\Tilde{h}}_{u}^{K}{-}\mathbf{h}_{u}^{K}||_{p} 
   \\ 
   \leq \prod_{k=1}^{K} ||\mathbf{W}_{a}^{k}||_{p}||(\mathbf{\Tilde{b}}_{u}{-}\mathbf{b}_{u})||_{p},
\end{aligned}
\end{eqnarray}
where $K$ is the last GNN layer. In the case of $p=2$, the Lipschitz constant in the above equation is equal to the product of the largest singular values (\ie spectral norm) of weight matrices $\mathbf{W}_{a}^{k}$ and can be approximated with a small number of iterations of the power method. We thus perform spectral normalization on the weights of each layer and use the normalized weights $\mathbf{\Tilde{W}}_{a}^{k}$ in the \textsc{Upd} step of each layer.

\xhdr{Theorem 2 (\nifty Counterfactual Fairness)}~\textit{Given a non-linear activation function $\sigma$ that is Lipschitz continuous and a binary valued sensitive attribute $s$, the (counterfactual) unfairness of the representations learned by our framework \nifty can be bounded as follows:
} 
\begin{equation}\label{eqn:theorem2}
    ||\enc(\Tilde{u}^s) - \enc(u)||_{p} \leq \prod_{k=1}^{K} ||\mathbf{W}_{a}^{k}||_{p} 
\end{equation}
\looseness=-1
where $\Tilde{u}^{s}$ is a node in the augmented graph which is generated by modifying (flipping) the value of the sensitive attribute (s) of node $u$ while keeping everything else constant.

\textit{Proof Sketch.~}
In order to prove this theorem, we will first prove the following: 
\begin{equation}\label{eqn:thm2sub}
    ||\enc(\Tilde{u}^s) - \enc(u)||_{p} \leq \prod_{k=1}^{K} ||\mathbf{W}_{a}^{k}||_{p} ||(\Tilde{\mathbf{b}}^{s}_{u}-\mathbf{b}_{u})||_{p} 
\end{equation}
It can be seen that the above equation has a similar form as that of Eqn.~\ref{eqn:theorem1} in Theorem 1. Therefore, the above equation can be proved analogously. Note that the node $\Tilde{u}^s$ in Eqn.~\ref{eqn:thm2sub} is exactly the same as the node $u$ except that the value of the sensitive attribute is flipped (either from 0 to 1, or from 1 to 0). Therefore, $||(\Tilde{\mathbf{b}}^{s}_{u}-\mathbf{b}_{u})||_{p} = 1$ and we obtain Eqn.~\ref{eqn:theorem2}. 
\xhdr{Proposition 1 (Counterfactual Fairness of Downstream Classifier)} \textit{If the representations learned by our framework \nifty satisfy counterfactual fairness, then a downstream classifier $f: \mathbf{z}_{u} \to \hat{y}_{u}$ which leverages these representations also satisfies counterfactual fairness.
}

Proof is provided in the Appendix~\ref{sec:app_proof_1}. 



\section{Experiments}
\label{sec:experim}
\begin{figure*}[t]
    \centering
    \vspace{-2mm}
    \begin{flushleft}
			\hspace{1.2cm}(a) German credit graph 
			\hspace{2.5cm}(b) Recidivism graph
			\hspace{2.1cm}(c) Credit defaulter graph
	\end{flushleft}
	\centering
	\includegraphics[width=0.95\textwidth]{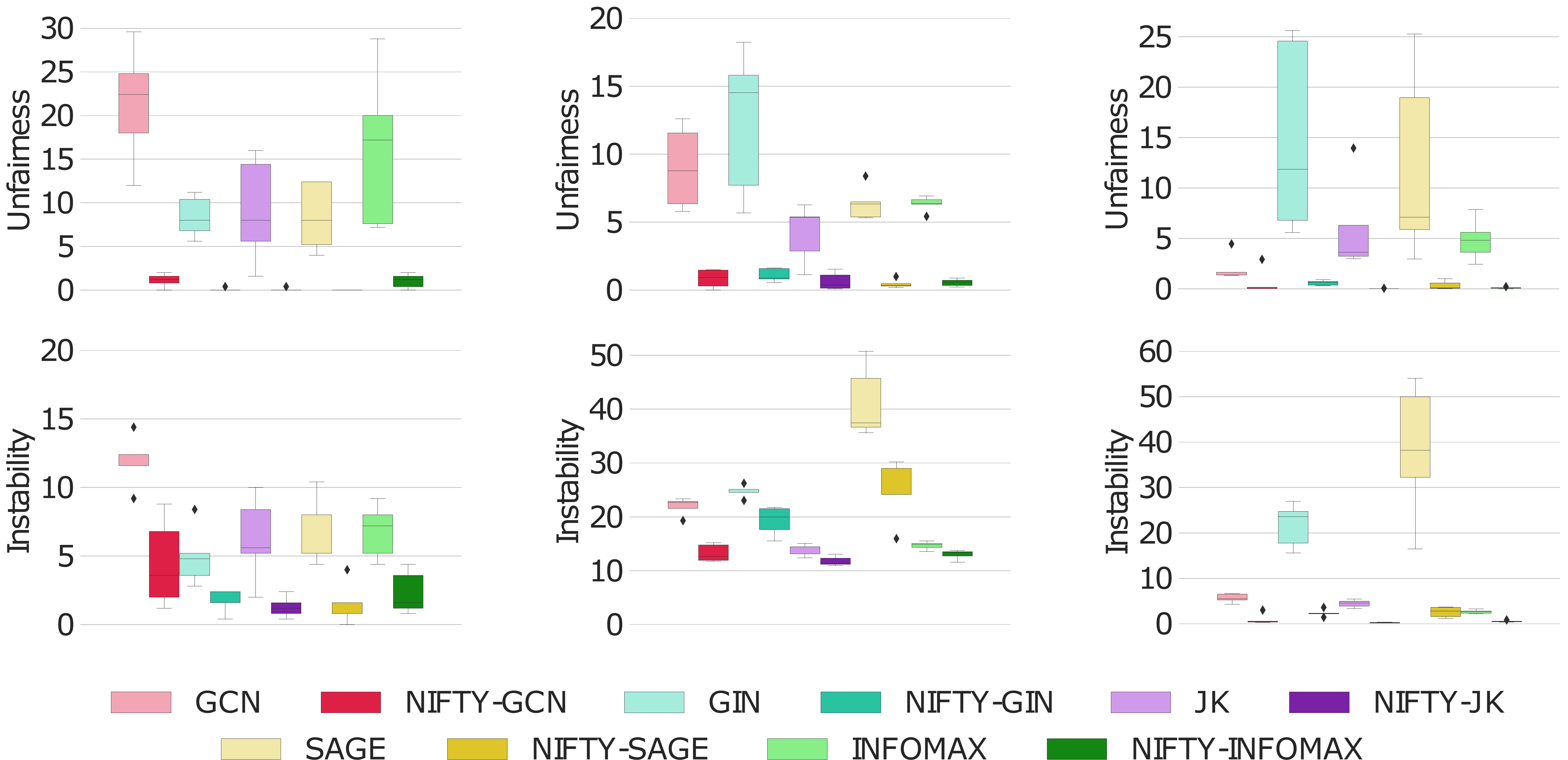}
	\caption{
	Unfairness (top) and instability (bottom) error rates for five GNNs and their \nifty counterparts.
	\nifty-enhanced GNNs give fairer and more stable predictions than their unmodified counterparts across all three datasets and five GNNs.}
    \label{fig:fairstab}
\end{figure*}

\begin{table*}
\setlength{\tabcolsep}{5pt}
\renewcommand{\arraystretch}{0.65}
\caption{Comparison of \nifty to existing methods for improving fairness (\ie FairGCN~\citep{dai2020fairgnn}) and stability (\ie RobustGCN~\citep{zhu2019robust}) of GNNs.
Shown is average performance across five independent runs. 
The counterfactual fairness does not apply to FairGCN (\ie N/A) as FairGCN cannot consider sensitive attributes.
Arrows ($\uparrow$, $\downarrow$) indicate the direction of better performance.
\nifty outperforms baselines methods by a large margin.
}
\centering
\label{tab:baseline}
{\small \begin{tabular}{llccaaaa}
Dataset & Method & AUROC ($\uparrow$) & F1-score ($\uparrow$) & \begin{tabular}[c]{@{}c@{}}Unfairness ($\downarrow$) \end{tabular} & \begin{tabular}[c]{@{}c@{}}Instability ($\downarrow$) \end{tabular} & \begin{tabular}[c]{@{}c@{}}$\Delta_{SP}$ ($\downarrow$)\end{tabular} & \begin{tabular}[c]{@{}c@{}}$\Delta_{EO}$ ($\downarrow$)\end{tabular} \\ \toprule
German credit graph & \begin{tabular}[c]{@{}l@{}}FairGCN\\ RobustGCN\\ \nifty-GCN\end{tabular} & \begin{tabular}[c]{@{}c@{}}{75.21}\std{0.36}\\ 71.06\std{1.48}\\ 70.32\std{4.42}\end{tabular} & \begin{tabular}[c]{@{}c@{}}81.52\std{0.68}\\ 78.85\std{6.39}\\ {81.98}\std{0.82}\end{tabular} & \begin{tabular}[c]{@{}c@{}}N/A\\ 7.68\std{4.69}\\ \textbf{1.12}\std{0.77}\end{tabular} & \begin{tabular}[c]{@{}c@{}}7.84\std{2.20}\\ 4.48\std{1.07}\\ \textbf{4.48}\std{3.23}\end{tabular} & \begin{tabular}[c]{@{}c@{}}38.12\std{4.87}\\ 25.78\std{10.92}\\ \textbf{15.08}\std{8.22}\end{tabular} & \begin{tabular}[c]{@{}c@{}}26.70\std{4.27}\\ 18.47\std{9.87}\\ \textbf{12.56}\std{8.60}\end{tabular} \\ \midrule
Recidivism graph & \begin{tabular}[c]{@{}l@{}}FairGCN\\ RobustGCN\\ \nifty-GCN\end{tabular} & \begin{tabular}[c]{@{}c@{}}{87.55}\std{0.60}\\ 87.25\std{1.67} \\ 81.40\std{0.89}\end{tabular} & \begin{tabular}[c]{@{}c@{}}78.14\std{0.94}\\{79.02}\std{2.84}\\ 69.24\std{0.70}\end{tabular} & \begin{tabular}[c]{@{}c@{}}N/A\\ 2.61\std{1.58} \\ \textbf{0.84}\std{0.68}\end{tabular} & \begin{tabular}[c]{@{}c@{}}24.37\std{2.33}\\ \textbf{13.02}\std{6.06}\\ 13.28\std{1.62}\end{tabular} & \begin{tabular}[c]{@{}c@{}}6.51\std{0.77}\\ 5.36\std{1.28}\\ \textbf{3.16}\std{0.60}\end{tabular} & \begin{tabular}[c]{@{}c@{}}4.51\std{1.10}\\ 4.20\std{1.88}\\ \textbf{2.99}\std{0.40}\end{tabular} \\ \midrule
Credit defaulter graph & \begin{tabular}[c]{@{}l@{}}FairGCN\\ RobustGCN\\ \nifty-GCN\end{tabular} & \begin{tabular}[c]{@{}c@{}}72.69\std{1.23}\\ {72.98}\std{0.26}\\ 71.92\std{0.19}\end{tabular} & \begin{tabular}[c]{@{}c@{}}80.16\std{2.03}\\ 81.79\std{0.60}\\ {81.99}\std{0.63}\end{tabular} & \begin{tabular}[c]{@{}c@{}}N/A\\ 0.94\std{0.60}\\ \textbf{0.63}\std{1.28}\end{tabular} & \begin{tabular}[c]{@{}c@{}}5.73\std{0.60}\\ 1.68\std{0.83}\\ \textbf{0.95}\std{1.16}\end{tabular} & \begin{tabular}[c]{@{}c@{}}15.86\std{5.16}\\ 12.41\std{0.54}\\ \textbf{12.40}\std{1.62}\end{tabular} & \begin{tabular}[c]{@{}c@{}}14.43\std{6.06}\\ 10.16\std{0.49}\\ \textbf{10.09}\std{1.55}\end{tabular}\\\bottomrule
\end{tabular}}
\vspace{-2mm}
\end{table*} 

Next, we present experimental results for our \nifty framework. We address the following key questions: Q1) Does \nifty enable GNNs to learn fair and stable embeddings? Q2) Can \nifty achieve group fairness? Q3) How does the interplay between fairness and stability affect downstream performance? Q4) Are changes to GNN's architecture and objective function necessary for fair and stable predictions?


\vspace{-2mm}
\subsection{Datasets and Experimental Setup}\label{sec:exp_setup}

We first describe datasets designed to study fair and stable network embeddings and then outline experimental setup. 

\xhdr{Datasets}
We construct three new datasets. 
1) The \textit{German credit graph} has 1,000 nodes representing clients in a German bank that are connected based on the similarity of their credit accounts. The task is to classify clients into good vs.~bad credit risks considering clients' gender as the sensitive attribute~\citep{Dua:2019}.
2) The \textit{Recidivism graph} has 18,876 nodes representing defendants who got released on bail at the U.S state courts during 1990-2009~\citep{jordan2015effect}. Defendants are connected based on the similarity of past criminal records and demographics. The goal is to classify defendants into bail (\ie unlikely to commit a violent crime if released) vs.~no bail (\ie likely to commit a violent crime) considering race information as the protected attribute.
3) The \textit{Credit defaulter graph} has 30,000 nodes representing individuals that we connected based on the similarity of their spending and payment patterns~\citep{yeh2009comparisons}. The task is to predict whether an individual will default on the credit card payment or not while considering age as the sensitive attribute. See Appendix for details on dataset construction.

\xhdr{Performance evaluation} 
To measure predictive performance of downstream binary node classification, we use AUROC and F1-score. To quantify group fairness, we use statistical parity (SP)~\citep{dwork12}, defined as: $\Delta_{SP}{=}|P(\hat{y}_{u}{=}1|s{=}0){-}P(\hat{y}_{u}{=}1|s{=}1)|$, and equal opportunity (EO)~\citep{hardt16}, defined as: $\Delta_{EO}{=}|P(\hat{y}_{u}{=}1|y_{u}{=}1,s{=}0){-}P(\hat{y}_{u}{=}1|y_{u}{=}1,s{=}1)|$, where probabilities are estimated on the test set~\citep{dai2020fairgnn}. To measure counterfactual fairness, we define the unfairness score as the percentage of test nodes for which predicted label changes when the node's sensitive attribute is flipped. 
Finally, the instability score represents the percentage of test nodes for which predicted label changes when random noise is added to node attributes. 



\xhdr{GNN methods} To investigate the flexibility of \nifty, we incorporate it into five  estabished and state-of-the-art GNN methods: GCN~\citep{kipf17:semi}, GraphSAGE~\citep{hamilton2017inductive}, Jumping Knowledge (JK)~\citep{xu2018representation}, GIN~\citep{xu2018powerful}, and InfoMax~\citep{velivckovic2018deep}.

\xhdr{Baseline methods and implementation} 
We consider two baseline methods: FairGCN \citep{dai2020fairgnn} and RobustGCN \citep{zhu2019robust}; all hyperparameters are set following the authors' guidelines. 
We use stop-gradient operation for training the Siamese networks~\citep{chen2020exploring}. We set regularization coefficient to $\lambda=0.6$ in all our experiments and conduct a sensitivity analysis into the effect of $\lambda$ on \nifty's performance. See Appendix for details.

\vspace{-2mm}
\subsection{Results}
\label{sec:fairstab}

\begin{table*}[t]
\centering
\caption{
Results of \nifty for five GNNs and three graph datasets.
Shown is average performance across five independent runs. Arrows ($\uparrow$, $\downarrow$) indicate the direction of better performance.
\nifty keeps the predictive power (AUROC and F1-score) of original GNNs while improving their fairness and stability (shaded area).
}
\label{tab:group_acc}
\setlength{\tabcolsep}{5pt}
\renewcommand{\arraystretch}{0.95}
{\small \begin{tabular}{clccaaaa}
Dataset & Method & AUROC ($\uparrow$) & F1-score ($\uparrow$) & Unfairness ($\downarrow$) & Instability ($\downarrow$) & $\Delta_{SP} (\downarrow$) & $\Delta_{EO} (\downarrow$) \\
\toprule
\multirow{10}{2cm}{German credit graph} & \begin{tabular}[c]{@{}l@{}}GCN\\\nifty-GCN\end{tabular} & \begin{tabular}[c]{@{}c@{}}{74.00}\std{1.51}\\ 70.32\std{4.42}\end{tabular} &
\begin{tabular}[c]{@{}c@{}}80.05\std{1.20} \\ {81.98}\std{0.82}\end{tabular} & 
\begin{tabular}[c]{@{}c@{}}21.36\std{6.70}\\ \textbf{1.12}\std{0.77}\end{tabular} & 
\begin{tabular}[c]{@{}c@{}}11.84\std{1.87} \\ \textbf{4.48}\std{3.23}\end{tabular} & \begin{tabular}[c]{@{}c@{}}41.94\std{5.52}\\ \textbf{15.08}\std{8.22} \end{tabular} & \begin{tabular}[c]{@{}c@{}}31.11\std{4.40}\\ \textbf{12.56}\std{8.60} \end{tabular} \\ 
 & \begin{tabular}[c]{@{}l@{}}GIN\\\nifty-GIN\end{tabular} & \begin{tabular}[c]{@{}c@{}}{72.69}\std{1.02}\\ 69.46\std{3.99}\end{tabular} & \begin{tabular}[c]{@{}c@{}}82.62\std{1.55} \\ {82.77}\std{0.48}\end{tabular} &
 \begin{tabular}[c]{@{}c@{}}8.40\std{2.37} \\ \textbf{0.08}\std{0.18}\end{tabular} &
 \begin{tabular}[c]{@{}c@{}}4.96\std{2.15} \\ \textbf{1.84}\std{0.88}\end{tabular} &
 \begin{tabular}[c]{@{}c@{}}14.85\std{4.64}\\ \textbf{4.39}\std{3.47}\end{tabular} & \begin{tabular}[c]{@{}c@{}}8.28\std{6.72}\\ \textbf{2.82}\std{1.60}\end{tabular} \\
 & \begin{tabular}[c]{@{}l@{}}GraphSAGE\\\nifty-GraphSAGE\end{tabular} & \begin{tabular}[c]{@{}c@{}}{74.54}\std{0.86}\\70.54\std{2.03}\end{tabular} & \begin{tabular}[c]{@{}c@{}}{81.15}\std{0.97}\\ 78.14\std{2.40}\end{tabular} &
 \begin{tabular}[c]{@{}c@{}}8.40\std{3.93}\\ \textbf{0.00}\std{0.00}\end{tabular} &
 \begin{tabular}[c]{@{}c@{}}6.64\std{2.51}\\ \textbf{1.44}\std{1.54}\end{tabular} &
 \begin{tabular}[c]{@{}c@{}}23.79\std{6.70}\\ \textbf{6.10}\std{4.93}\end{tabular} & \begin{tabular}[c]{@{}c@{}}15.13\std{5.74}\\ \textbf{6.34}\std{3.57}\end{tabular} \\
 & \begin{tabular}[c]{@{}l@{}}Infomax\\\nifty-Infomax\end{tabular} & \begin{tabular}[c]{@{}c@{}}67.98\std{3.94}\\ {72.01}\std{2.05}\end{tabular} & \begin{tabular}[c]{@{}c@{}}72.70\std{7.91}\\ {81.98}\std{0.33}\end{tabular} &
 \begin{tabular}[c]{@{}c@{}}16.16\std{9.07}\\ \textbf{1.04}\std{0.83}\end{tabular} &
 \begin{tabular}[c]{@{}c@{}}6.80\std{1.98}\\ \textbf{2.32}\std{1.58}\end{tabular} &
 \begin{tabular}[c]{@{}c@{}}36.79\std{6.58}\\ \textbf{9.25}\std{6.45}\end{tabular} & \begin{tabular}[c]{@{}c@{}}28.99\std{5.70}\\ \textbf{7.21}\std{4.49}\end{tabular} \\
 & \begin{tabular}[c]{@{}l@{}}JK\\ \nifty-JK\end{tabular} & \begin{tabular}[c]{@{}c@{}}{71.49}\std{2.64}\\ 70.42\std{2.03}\end{tabular} & \begin{tabular}[c]{@{}c@{}}80.88\std{1.02}\\ {81.25}\std{0.93}\end{tabular} &
 \begin{tabular}[c]{@{}c@{}}9.12\std{6.03}\\ \textbf{0.08}\std{0.18}\end{tabular} &
 \begin{tabular}[c]{@{}c@{}}6.24\std{3.09}\\ \textbf{1.28}\std{0.77}\end{tabular} &
 \begin{tabular}[c]{@{}c@{}}20.12\std{5.16}\\ \textbf{4.98}\std{6.36}\end{tabular} & \begin{tabular}[c]{@{}c@{}}9.75\std{4.73}\\ \textbf{3.42}\std{3.52}\end{tabular} \\
 \midrule
\multirow{10}{2cm}{Recidivism graph} & \begin{tabular}[c]{@{}l@{}}GCN\\ \nifty-GCN\end{tabular} & \begin{tabular}[c]{@{}c@{}}{86.52}\std{0.42}\\ 81.40\std{0.89}\end{tabular} & \begin{tabular}[c]{@{}c@{}}{77.50}\std{0.87}\\ 69.24\std{0.70}\end{tabular} &
\begin{tabular}[c]{@{}c@{}}9.02\std{3.04}\\ \textbf{0.84}\std{0.68}\end{tabular} &
\begin{tabular}[c]{@{}c@{}}21.97\std{1.63}\\ \textbf{13.28}\std{1.62}\end{tabular} &
\begin{tabular}[c]{@{}c@{}}8.49\std{0.73}\\ \textbf{3.16}\std{0.60}\end{tabular} & \begin{tabular}[c]{@{}c@{}}5.93\std{0.56}\\ \textbf{2.99}\std{0.40}\end{tabular} \\
 & \begin{tabular}[c]{@{}l@{}}GIN\\ \nifty-GIN\end{tabular} & \begin{tabular}[c]{@{}c@{}}81.32\std{1.61}\\ {84.28}\std{1.42}\end{tabular} & \begin{tabular}[c]{@{}c@{}}70.97\std{2.48}\\ {72.07}\std{6.14}\end{tabular} &
 \begin{tabular}[c]{@{}c@{}}12.40\std{5.42}\\ \textbf{1.09}\std{0.49}\end{tabular} &
 \begin{tabular}[c]{@{}c@{}}24.82\std{1.16}\\ \textbf{19.29}\std{2.67}\end{tabular} &
 \begin{tabular}[c]{@{}c@{}}9.91\std{3.24}\\ \textbf{6.57}\std{1.77}\end{tabular} & \begin{tabular}[c]{@{}c@{}}6.83\std{3.02}\\ \textbf{5.17}\std{2.15}\end{tabular} \\
 & \begin{tabular}[c]{@{}l@{}}GraphSAGE\\ \nifty-GraphSAGE\end{tabular} & \begin{tabular}[c]{@{}c@{}}91.29\std{0.95}\\ {92.43}\std{0.44}\end{tabular} & \begin{tabular}[c]{@{}c@{}}81.58\std{1.52}\\ {82.08}\std{2.40}\end{tabular} &
 \begin{tabular}[c]{@{}c@{}}6.39\std{1.24}\\ \textbf{0.46}\std{0.32}\end{tabular} &
 \begin{tabular}[c]{@{}c@{}}41.24\std{6.67}\\ \textbf{25.66}\std{5.90}\end{tabular} &
 \begin{tabular}[c]{@{}c@{}}\textbf{1.82}\std{1.51}\\ 6.43\std{0.67}\end{tabular} & \begin{tabular}[c]{@{}c@{}}\textbf{2.16}\std{0.24}\\ 5.23\std{1.26}\end{tabular} \\
 & \begin{tabular}[c]{@{}l@{}}Infomax\\ \nifty-Infomax\end{tabular} & \begin{tabular}[c]{@{}c@{}}{89.24}\std{0.08}\\ 79.67\std{0.44}\end{tabular} & \begin{tabular}[c]{@{}c@{}}{80.11}\std{0.16}\\ 67.77\std{1.47}\end{tabular} &
 \begin{tabular}[c]{@{}c@{}}6.34\std{0.57}\\ \textbf{0.56}\std{0.27}\end{tabular} &
 \begin{tabular}[c]{@{}c@{}}14.69\std{0.75}\\ \textbf{13.03}\std{0.88}\end{tabular} &
 \begin{tabular}[c]{@{}c@{}}7.41\std{0.48}\\ \textbf{4.04}\std{0.24}\end{tabular} & \begin{tabular}[c]{@{}c@{}}\textbf{3.04}\std{0.46}\\ 3.43\std{0.38}\end{tabular} \\
 & \begin{tabular}[c]{@{}l@{}}JK\\ \nifty-JK\end{tabular} & \begin{tabular}[c]{@{}c@{}}{88.60}\std{0.45}\\ 81.73\std{0.38}\end{tabular} & \begin{tabular}[c]{@{}c@{}}{79.61}\std{0.82}\\ 70.20\std{1.20}\end{tabular} &
 \begin{tabular}[c]{@{}c@{}}4.20\std{2.14}\\ \textbf{0.64}\std{0.65}\end{tabular} &
 \begin{tabular}[c]{@{}c@{}}13.64\std{1.09}\\ \textbf{11.79}\std{0.88}\end{tabular} &
 \begin{tabular}[c]{@{}c@{}}7.60\std{0.71}\\ \textbf{4.28}\std{1.17}\end{tabular} & \begin{tabular}[c]{@{}c@{}}4.25\std{0.25}\\ \textbf{3.65}\std{1.03}\end{tabular} \\
 \midrule
\multirow{10}{2.1cm}{Credit defaulter graph} & \begin{tabular}[c]{@{}l@{}}GCN\\ \nifty-GCN\end{tabular} & \begin{tabular}[c]{@{}c@{}}{72.97}\std{1.63}\\ 71.92\std{0.19}\end{tabular} & \begin{tabular}[c]{@{}c@{}}{82.02}\std{0.45}\\ 81.99\std{0.63}\end{tabular} &
\begin{tabular}[c]{@{}c@{}}2.04\std{1.36}\\ \textbf{0.63}\std{1.28}\end{tabular} &
\begin{tabular}[c]{@{}c@{}}5.63\std{0.98}\\ \textbf{0.95}\std{1.16}\end{tabular} &
\begin{tabular}[c]{@{}c@{}}\textbf{10.76}\std{5.21}\\ 12.40\std{1.62}\end{tabular} & \begin{tabular}[c]{@{}c@{}}\textbf{8.71}\std{4.81}\\ 10.09\std{1.55}\end{tabular} \\
 & \begin{tabular}[c]{@{}l@{}}GIN\\ \nifty-GIN\end{tabular} & \begin{tabular}[c]{@{}c@{}}{73.71}\std{0.33}\\ 71.28\std{0.19}\end{tabular} & \begin{tabular}[c]{@{}c@{}}82.04\std{0.60}\\ {84.97}\std{0.58}\end{tabular} &
 \begin{tabular}[c]{@{}c@{}}14.89\std{9.63}\\ \textbf{0.59}\std{0.24}\end{tabular} &
 \begin{tabular}[c]{@{}c@{}}21.73\std{4.81}\\ \textbf{2.36}\std{0.78}\end{tabular} &
 \begin{tabular}[c]{@{}c@{}}13.48\std{2.45}\\ \textbf{4.93}\std{3.75}\end{tabular} & \begin{tabular}[c]{@{}c@{}}11.19\std{3.20}\\ \textbf{4.60}\std{2.80}\end{tabular} \\
 & \begin{tabular}[c]{@{}l@{}}GraphSAGE\\ \nifty-GraphSAGE\end{tabular} & \begin{tabular}[c]{@{}c@{}}{75.19}\std{0.15}\\ 73.27\std{0.21}\end{tabular} & \begin{tabular}[c]{@{}c@{}}82.78\std{0.37}\\ {83.64}\std{1.66}\end{tabular} &
 \begin{tabular}[c]{@{}c@{}}12.04\std{9.60}\\ \textbf{0.35}\std{0.44}\end{tabular} &
 \begin{tabular}[c]{@{}c@{}}38.19\std{14.97}\\ \textbf{2.57}\std{1.15}\end{tabular} &
 \begin{tabular}[c]{@{}c@{}}15.66\std{1.62}\\ \textbf{12.65}\std{0.95}\end{tabular} & \begin{tabular}[c]{@{}c@{}}13.52\std{1.47}\\ \textbf{9.93}\std{0.67}\end{tabular} \\
 & \begin{tabular}[c]{@{}l@{}}Infomax\\ \nifty-Infomax\end{tabular} & \begin{tabular}[c]{@{}c@{}}{74.17}\std{0.11}\\ 71.86\std{0.26}\end{tabular} & \begin{tabular}[c]{@{}c@{}}{82.58}\std{0.33}\\ 81.70\std{0.06}\end{tabular} &
 \begin{tabular}[c]{@{}c@{}}4.87\std{2.07}\\ \textbf{0.09}\std{0.08}\end{tabular} &
 \begin{tabular}[c]{@{}c@{}}2.67\std{0.43}\\ \textbf{0.53}\std{0.20}\end{tabular} &
 \begin{tabular}[c]{@{}c@{}}14.57\std{0.69}\\ \textbf{11.83}\std{0.36}\end{tabular} & \begin{tabular}[c]{@{}c@{}}12.26\std{0.72}\\ \textbf{9.52}\std{0.31}\end{tabular} \\
 & \begin{tabular}[c]{@{}l@{}}JK\\ \nifty-JK\end{tabular} & \begin{tabular}[c]{@{}c@{}}{73.80}\std{0.06}\\ 72.07\std{0.30}\end{tabular} & \begin{tabular}[c]{@{}c@{}}{82.70}\std{0.73}\\ 81.78\std{0.08}\end{tabular} &
 \begin{tabular}[c]{@{}c@{}}6.03\std{4.63}\\ \textbf{0.02}\std{0.02}\end{tabular} &
 \begin{tabular}[c]{@{}c@{}}4.45\std{0.83}\\ \textbf{0.26}\std{0.09}\end{tabular} &
 \begin{tabular}[c]{@{}c@{}}12.70\std{1.74}\\ \textbf{11.77}\std{0.09}\end{tabular} & \begin{tabular}[c]{@{}c@{}}9.51\std{0.07}\\ \textbf{9.42}\std{0.37}\end{tabular}\\\bottomrule
\end{tabular}}
\end{table*}

Next, we discuss experimental results that answer key questions highlighted at the beginning of this section (Q1-Q4). 



\xhdr{Q1) \nifty improves fairness and stability of GNNs}
Across three datasets and five GNNs, Fig.~\ref{fig:fairstab} shows that \nifty-augmented GNNs learn fairer and more stable embeddings than unmodified GNNs. 
On average, \nifty improves stability and fairness of GNNs by $60.87\%$ and $92.01\%$, respectively.
Further, \nifty can promote fairness and stability of GNNs without sacrificing their predictive performance, as evidenced by AUROC and F1-scores in Table~\ref{tab:group_acc}.
Finally, \nifty outperforms baseline FairGCN and RobustGCN methods by $62.07\%$ and $57.26\%$ on four fairness and stability metrics (Table~\ref{tab:baseline}).
\xhdr{Q2) \nifty achieves group fairness} 
Remarkably, while \nifty's explicit aim is to capture counterfactual fairness, our approach indirectly improves group
fairness of GNNs because it reduces information on protected attributes, and, we argue, makes the multi-objective problem of satisfying fairness and stability more tractable. Across three datasets, five GNNs, and two group fairness metrics, \nifty achieves $43.56\%$ lower $\Delta_{SP}$ and $34.70\%$ lower $\Delta_{EO}$.
Further, we find that \nifty achieves $36.05\%$ lower $\Delta_{SP}$ and $29.71\%$ lower $\Delta_{EO}$ error rates than baseline methods (Table~\ref{tab:baseline}), suggesting that in \nifty, a node's chance of being represented as a particular point in the embedding space does not depend on the node's membership in a protected group.
\begin{figure}
    \centering
    \includegraphics[width=0.45\textwidth]{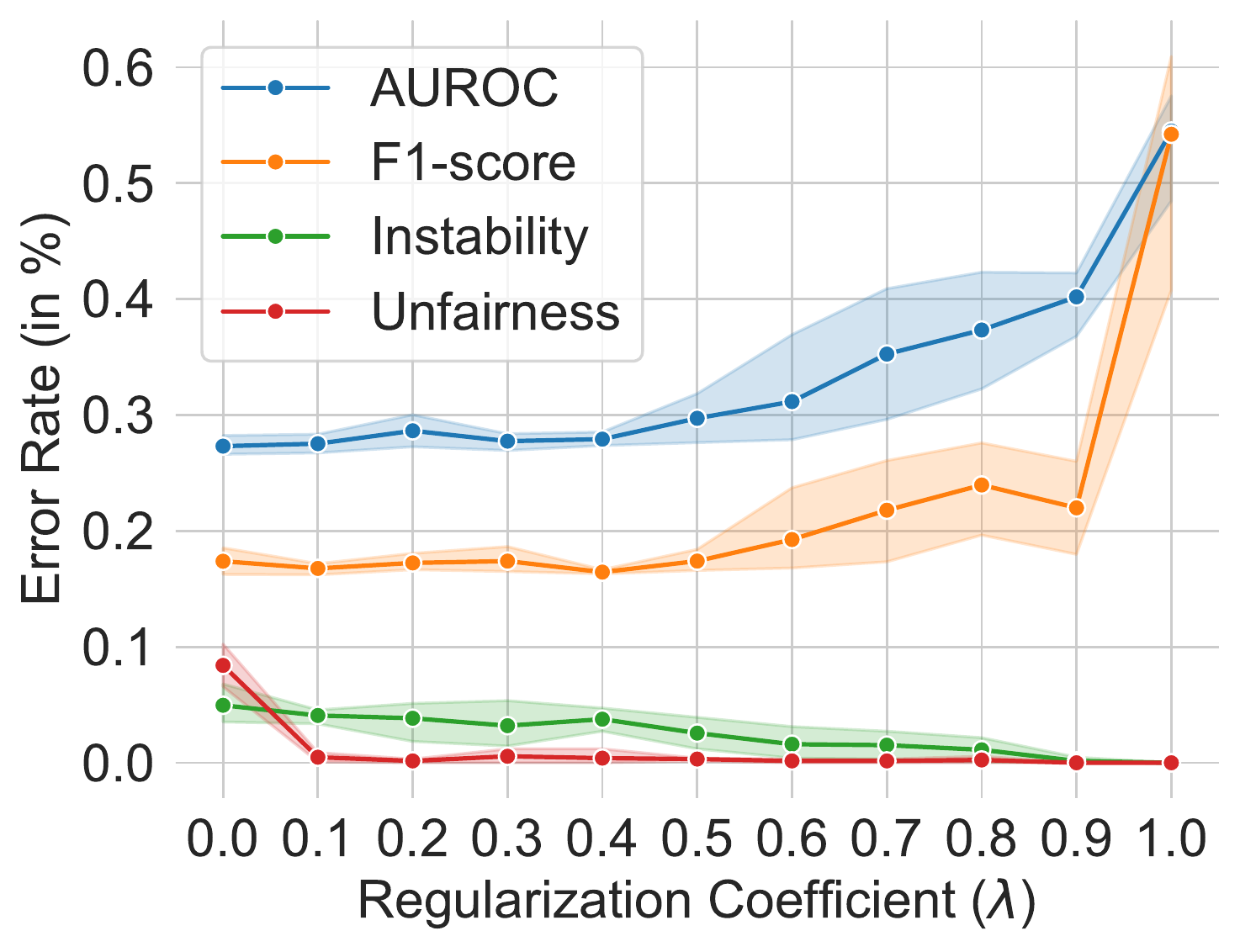}    
    \caption{
    The effects of regularization on the performance of \nifty. 
    Shown are results for \nifty-GIN and the German credit graph (see Fig.~\ref{fig:app_tradeoff} for other datasets).
    Over a wide range of regularization strength ($0.1 < \lambda < 0.5$), \nifty achieves a near-perfect stability and fairness on the downstream task without sacrificing the predictive ability of GIN.
    }
    \label{fig:tradeoff}
\end{figure}

\xhdr{Q3) Trade-offs between fairness, stability, and predictive performance}
As we increase regularization coefficient $\lambda$ in \nifty (Fig.~\ref{fig:tradeoff}), we find that the error rates for counterfactual fairness and stability steadily decrease.
Interestingly, even with a modest amount of regularization ($\lambda=0.1$), \nifty achieves a $94.29\%$ improvement in unfairness error rate.
As expected, a more strongly regularized \nifty model takes a hit on its predictive performance (higher error rate for AUROC and F1-score).
See Fig.~\ref{fig:app_tradeoff} for similar trends on the recidivism and credit defaulter graphs.
\hide{
We observe a steady decrease in the error rate for counterfactual fairness and stability as we increase the regularization value (Fig.~\ref{fig:tradeoff}).
Interestingly, even with a small regularization of $\lambda=0.1$, our model achieves a $94.29\%$ improvement in unfairness error rate.
For higher regularization values, the model takes a hit on its precision performance (higher error rate for AUROC and F1-score).
Here, for the first time, we show the trade-off between the accuracy (AUROC and F1-score), fairness, and stability of a model.
See Fig.~\ref{fig:app_tradeoff} for similar trends for recidivism and credit datasets.
}
\begin{table*}[t]
\setlength{\tabcolsep}{5pt}
\centering
\caption{
Ablation study on the recidivism graph. Shown is average performance across five independent runs, evidencing that \nifty's changes in the GNN architecture and the objective function are complementary and improve fairness and stability. 
}
\label{tab:ablation}
\vspace{-2mm}
{\small \begin{tabular}{lccaaaa}
\multicolumn{1}{c}{Method} & AUROC ($\uparrow$) & F1-score ($\uparrow$) & Unfairness ($\downarrow$) & Instability ($\downarrow$) & \begin{tabular}[c]{@{}c@{}}$\Delta_{SP} (\downarrow)$\end{tabular} & \begin{tabular}[c]{@{}c@{}}$\Delta_{EO} (\downarrow)$\end{tabular} \\\toprule
\begin{tabular}[c]{@{}l@{}}GCN~\citep{kipf17:semi}\\ \nifty-GCN w/o obj. changes (Sec.~\ref{sec:objective})\\ \nifty-GCN w/o arch. changes (Sec.~\ref{sec:architecture})\\ \nifty-GCN\end{tabular} & \begin{tabular}[c]{@{}c@{}}86.52\std{0.42}\\ 80.02 \std{0.20}\\ 84.83 \std{2.85}\\ 81.40 \std{0.89}\end{tabular} & \begin{tabular}[c]{@{}c@{}}77.50\std{0.87}\\ 67.51 \std{0.23}\\ 76.15\std{5.74}\\ 69.24\std{0.70}\end{tabular} & \begin{tabular}[c]{@{}c@{}}9.02\std{3.04}\\ 2.61\std{0.64}\\ 1.64 \std{1.58}\\ \textbf{0.84}\std{0.68}\end{tabular} & \begin{tabular}[c]{@{}c@{}}21.97\std{1.63}\\ 13.69\std{0.60}\\ 13.98 \std{1.38}\\ \textbf{13.28}\std{1.62}\end{tabular} & \begin{tabular}[c]{@{}c@{}}8.49\std{0.73}\\ 5.86\std{0.85}\\ 4.29 \std{1.32}\\ \textbf{3.16}\std{0.60}\end{tabular} & \begin{tabular}[c]{@{}c@{}}5.93\std{0.56}\\ 4.65\std{0.49}\\ 3.48 \std{1.37}\\ \textbf{2.99}\std{0.40}\end{tabular} \\\bottomrule
\end{tabular}}
\vspace{-2mm}
\end{table*}

\xhdr{Q4) Ablation study}
We conduct ablations on two key \nifty's components, namely the objective function and the layer-wise normalization of GNN's architecture using the Lipschitz constant.
Results show that both components are necessary to generate embeddings that are simultaneously fair and stable (Table~\ref{tab:ablation}).
In particular, we observe a $90.7\%$ improvement in fairness of \nifty-GCN as compared to vanilla GCN, providing empirical evidence for our theoretical analysis that the Lipschitz normalization can improve both fairness and stability of graph embeddings (Section~\ref{sec:theory}).

\section{Conclusions \& Future Work}
\label{sec:conclusion}
We propose and address the problem of learning representations that are both fair and stable. 
To this end, we introduce \nifty, a unified framework which exploits a key connection between counterfactual fairness and stability to learn representations that satisfy both these properties. 
At its core, \nifty, outlines a two-level strategy to modify an existing GNN both at the architectural as well as the objective function level. We carry out detailed theoretical analysis to show that the representations learned by \nifty are both counterfactually fair and stable. 
Further, results on new graph datasets from domains such as criminal justice and financial lending show that \nifty can considerably improve fairness (both in terms of counterfactual and group fairness) and stability without sacrificing predictive performance. This work paves way for several exciting future directions. For instance, it would be interesting to extend \nifty to generate fair and stable representations of other graph components (\eg edges, subgraphs) and to cater to other downstream tasks (\eg link prediction, graph classification). 

\begin{acknowledgements}
    We would like to thank the anonymous reviewers for their insightful feedback. H.L. is supported, in part, by the NSF award IIS-2008461, and Google. M.Z. is supported, in part, by NSF under nos. IIS-2030459 and IIS-2033384, the Harvard Data Science Initiative, the Amazon Research Award, and the Bayer Early Excellence in Science Award. The views expressed are those of the authors and do not reflect the official policy or position of the funding agencies.
\end{acknowledgements}

\bibliography{refs}

\begin{thebibliography}{44}
\providecommand{\natexlab}[1]{#1}
\providecommand{\url}[1]{\texttt{#1}}
\expandafter\ifx\csname urlstyle\endcsname\relax
  \providecommand{\doi}[1]{doi: #1}\else
  \providecommand{\doi}{doi: \begingroup \urlstyle{rm}\Url}\fi

\bibitem[Alsentzer et~al.(2020)Alsentzer, Finlayson, Li, and
  Zitnik]{alsentzer2020subgraph}
Emily Alsentzer, Samuel~G Finlayson, Michelle~M Li, and Marinka Zitnik.
\newblock Subgraph neural networks.
\newblock In \emph{NeurIPS}, 2020.

\bibitem[Berk et~al.(2018)Berk, Heidari, Jabbari, Kearns, and Roth]{BerkHJKR18}
Richard Berk, Hoda Heidari, Shahin Jabbari, Michael Kearns, and Aaron Roth.
\newblock Fairness in criminal justice risk assessments: {T}he state of the
  art.
\newblock In \emph{Sociological Methods \& Research}, 2018.

\bibitem[Bose and Hamilton(2019)]{bose2019compositional}
Avishek~Joey Bose and William~L Hamilton.
\newblock Compositional fairness constraints for graph embeddings.
\newblock In \emph{ICML}, 2019.

\bibitem[Bromley et~al.(1994)Bromley, Guyon, LeCun, S{\"a}ckinger, and
  Shah]{bromley1994signature}
Jane Bromley, Isabelle Guyon, Yann LeCun, Eduard S{\"a}ckinger, and Roopak
  Shah.
\newblock Signature verification using a" siamese" time delay neural network.
\newblock In \emph{NeurIPS}, 1994.

\bibitem[Chen et~al.(2020)Chen, Kornblith, Norouzi, and Hinton]{chen2020simple}
Ting Chen, Simon Kornblith, Mohammad Norouzi, and Geoffrey Hinton.
\newblock A simple framework for contrastive learning of visual
  representations.
\newblock In \emph{ICML}, 2020.

\bibitem[Chen and He(2020)]{chen2020exploring}
Xinlei Chen and Kaiming He.
\newblock Exploring simple siamese representation learning.
\newblock \emph{arXiv}, 2020.

\bibitem[Chouldechova(2017)]{Chouldechova17}
Alexandra Chouldechova.
\newblock Fair prediction with disparate impact: {A} study of bias in
  recidivism prediction instruments.
\newblock In \emph{Big Data}, 2017.

\bibitem[Dai and Wang(2021)]{dai2020fairgnn}
Enyan Dai and Suhang Wang.
\newblock Fairgnn: Eliminating the discrimination in graph neural networks with
  limited sensitive attribute information.
\newblock In \emph{WSDM}, 2021.

\bibitem[Defferrard et~al.(2016)Defferrard, Bresson, and
  Vandergheynst]{defferrard2016convolutional}
Micha{\"e}l Defferrard, Xavier Bresson, and Pierre Vandergheynst.
\newblock Convolutional neural networks on graphs with fast localized spectral
  filtering.
\newblock In \emph{NeurIPS}, 2016.

\bibitem[Dua and Graff(2017)]{Dua:2019}
Dheeru Dua and Casey Graff.
\newblock {UCI} machine learning repository, 2017.

\bibitem[Dwork et~al.(2012)Dwork, Hardt, Pitassi, Reingold, and Zemel]{dwork12}
Cynthia Dwork, Moritz Hardt, Toniann Pitassi, Omer Reingold, and Richard Zemel.
\newblock Fairness through awareness.
\newblock In \emph{ITCS}, 2012.

\bibitem[Fisher et~al.(2020)Fisher, Mittal, Palfrey, and
  Christodoulopoulos]{fisher2020debiasing}
Joseph Fisher, Arpit Mittal, Dave Palfrey, and Christos Christodoulopoulos.
\newblock Debiasing knowledge graph embeddings.
\newblock In \emph{EMNLP}, 2020.

\bibitem[Gainza et~al.(2020)Gainza, Sverrisson, Monti, Rodola, Boscaini,
  Bronstein, and Correia]{gainza2020deciphering}
Pablo Gainza, Freyr Sverrisson, Frederico Monti, Emanuele Rodola, D~Boscaini,
  MM~Bronstein, and BE~Correia.
\newblock Deciphering interaction fingerprints from protein molecular surfaces
  using geometric deep learning.
\newblock In \emph{Nature Methods}, 2020.

\bibitem[Geisler et~al.(2020)Geisler, Z{\"u}gner, and
  G{\"u}nnemann]{geisler2020reliable}
Simon Geisler, Daniel Z{\"u}gner, and Stephan G{\"u}nnemann.
\newblock Reliable graph neural networks via robust aggregation.
\newblock In \emph{NeurIPS}, 2020.

\bibitem[Gilmer et~al.(2017)Gilmer, Schoenholz, Riley, Vinyals, and
  Dahl]{gilmer2017neural}
Justin Gilmer, Samuel~S Schoenholz, Patrick~F Riley, Oriol Vinyals, and
  George~E Dahl.
\newblock Neural message passing for quantum chemistry.
\newblock In \emph{ICML}, 2017.

\bibitem[Gouk et~al.(2021)Gouk, Frank, Pfahringer, and
  Cree]{gouk2021regularisation}
Henry Gouk, Eibe Frank, Bernhard Pfahringer, and Michael~J Cree.
\newblock Regularisation of neural networks by enforcing lipschitz continuity.
\newblock In \emph{Machine Learning}. Springer, 2021.

\bibitem[Grill et~al.(2020)Grill, Strub, Altch{\'e}, Tallec, Richemond,
  Buchatskaya, Doersch, Pires, Guo, Azar, et~al.]{grill2020bootstrap}
Jean-Bastien Grill, Florian Strub, Florent Altch{\'e}, Corentin Tallec,
  Pierre~H Richemond, Elena Buchatskaya, Carl Doersch, Bernardo~Avila Pires,
  Zhaohan~Daniel Guo, Mohammad~Gheshlaghi Azar, et~al.
\newblock Bootstrap your own latent: A new approach to self-supervised
  learning.
\newblock In \emph{NeurIPS}, 2020.

\bibitem[Gysi et~al.(2020)Gysi, Do~Valle, Zitnik, Ameli, Gan, Varol, Sanchez,
  Baron, Ghiassian, Loscalzo, et~al.]{gysi2020network}
Deisy~Morselli Gysi, {\'I}talo Do~Valle, Marinka Zitnik, Asher Ameli, Xiao Gan,
  Onur Varol, Helia Sanchez, Rebecca~Marlene Baron, Dina Ghiassian, Joseph
  Loscalzo, et~al.
\newblock Network medicine framework for identifying drug repurposing
  opportunities for {COVID-19}.
\newblock \emph{arXiv}, 2020.

\bibitem[Hamilton et~al.(2017)Hamilton, Ying, and
  Leskovec]{hamilton2017inductive}
Will Hamilton, Zhitao Ying, and Jure Leskovec.
\newblock Inductive representation learning on large graphs.
\newblock In \emph{NeurIPS}, 2017.

\bibitem[Hammond et~al.(2011)Hammond, Vandergheynst, and
  Gribonval]{hammond2011wavelets}
David~K Hammond, Pierre Vandergheynst, and R{\'e}mi Gribonval.
\newblock Wavelets on graphs via spectral graph theory.
\newblock In \emph{Applied and Computational Harmonic Analysis}, 2011.

\bibitem[Hardt et~al.(2016)Hardt, Price, and Srebro]{hardt16}
Moritz Hardt, Eric Price, and Nathan Srebro.
\newblock Equality of opportunity in supervised learning.
\newblock In \emph{NeurIPS}, 2016.

\bibitem[Hu et~al.(2020)Hu, Dong, Wang, and Sun]{hu2020heterogeneous}
Ziniu Hu, Yuxiao Dong, Kuansan Wang, and Yizhou Sun.
\newblock Heterogeneous graph transformer.
\newblock In \emph{WWW}, 2020.

\bibitem[Huang et~al.(2020)Huang, Xiao, Glass, Zitnik, and
  Sun]{huang2020skipgnn}
Kexin Huang, Cao Xiao, Lucas~M Glass, Marinka Zitnik, and Jimeng Sun.
\newblock Skipgnn: predicting molecular interactions with skip-graph networks.
\newblock In \emph{Scientific Reports}, 2020.

\bibitem[Jin et~al.(2020)Jin, Wang, Zhu, Feng, Huang, and
  Zhou]{jin2020addressing}
Guangyin Jin, Qi~Wang, Cunchao Zhu, Yanghe Feng, Jincai Huang, and Jiangping
  Zhou.
\newblock Addressing crime situation forecasting task with temporal graph
  convolutional neural network approach.
\newblock In \emph{ICMTMA}, 2020.

\bibitem[Jordan and Freiburger(2015)]{jordan2015effect}
Kareem~L Jordan and Tina~L Freiburger.
\newblock The effect of race/ethnicity on sentencing: Examining sentence type,
  jail length, and prison length.
\newblock In \emph{Journal of Ethnicity in Criminal Justice}. Taylor \&
  Francis, 2015.

\bibitem[Kipf and Welling(2017)]{kipf17:semi}
Thomas~N Kipf and Max Welling.
\newblock Semi-supervised classification with graph convolutional networks.
\newblock In \emph{ICLR}, 2017.

\bibitem[Kleinberg et~al.(2017)Kleinberg, Mullainathan, and
  Raghavan]{KleinbergMR17}
Jon Kleinberg, Sendhil Mullainathan, and Manish Raghavan.
\newblock Inherent trade-offs in the fair determination of risk scores.
\newblock In \emph{ITCS}, 2017.

\bibitem[Kusner et~al.(2017)Kusner, Loftus, Russell, and
  Silva]{kusner2017counterfactual}
Matt~J Kusner, Joshua Loftus, Chris Russell, and Ricardo Silva.
\newblock Counterfactual fairness.
\newblock In \emph{NeurIPS}, 2017.

\bibitem[Lee et~al.(2019)Lee, Lee, and Kang]{lee2019self}
Junhyun Lee, Inyeop Lee, and Jaewoo Kang.
\newblock Self-attention graph pooling.
\newblock In \emph{ICML}, 2019.

\bibitem[Liao et~al.(2019)Liao, Huang, Kairouz, and Sankar]{liao2019learning}
Jiachun Liao, Chong Huang, Peter Kairouz, and Lalitha Sankar.
\newblock Learning generative adversarial representations (gap) under fairness
  and censoring constraints.
\newblock \emph{arXiv}, 2019.

\bibitem[Rahman et~al.(2019)Rahman, Surma, Backes, and
  Zhang]{rahman2019fairwalk}
Tahleen~A Rahman, Bartlomiej Surma, Michael Backes, and Yang Zhang.
\newblock Fairwalk: Towards fair graph embedding.
\newblock In \emph{IJCAI}, 2019.

\bibitem[Ustun et~al.(2019)Ustun, Spangher, and Liu]{ustun2019actionable}
Berk Ustun, Alexander Spangher, and Yang Liu.
\newblock Actionable recourse in linear classification.
\newblock In \emph{FAT}, 2019.

\bibitem[Veli{\v{c}}kovi{\'c} et~al.(2019)Veli{\v{c}}kovi{\'c}, Fedus,
  Hamilton, Li{\`o}, Bengio, and Hjelm]{velivckovic2018deep}
Petar Veli{\v{c}}kovi{\'c}, William Fedus, William~L Hamilton, Pietro Li{\`o},
  Yoshua Bengio, and R~Devon Hjelm.
\newblock Deep graph infomax.
\newblock In \emph{ICLR}, 2019.

\bibitem[Wu et~al.(2020)Wu, Pan, Chen, Long, Zhang, and
  Philip]{wu2020comprehensive}
Zonghan Wu, Shirui Pan, Fengwen Chen, Guodong Long, Chengqi Zhang, and S~Yu
  Philip.
\newblock A comprehensive survey on graph neural networks.
\newblock In \emph{IEEE Transactions on Neural Networks and Learning Systems},
  2020.

\bibitem[Xu et~al.(2018)Xu, Li, Tian, Sonobe, Kawarabayashi, and
  Jegelka]{xu2018representation}
Keyulu Xu, Chengtao Li, Yonglong Tian, Tomohiro Sonobe, Ken-ichi Kawarabayashi,
  and Stefanie Jegelka.
\newblock Representation learning on graphs with jumping knowledge networks.
\newblock In \emph{ICML}, 2018.

\bibitem[Xu et~al.(2019)Xu, Hu, Leskovec, and Jegelka]{xu2018powerful}
Keyulu Xu, Weihua Hu, Jure Leskovec, and Stefanie Jegelka.
\newblock How powerful are graph neural networks?
\newblock In \emph{ICLR}, 2019.

\bibitem[Yeh and Lien(2009)]{yeh2009comparisons}
I-Cheng Yeh and Che-hui Lien.
\newblock The comparisons of data mining techniques for the predictive accuracy
  of probability of default of credit card clients.
\newblock In \emph{ESA}, 2009.

\bibitem[Ying et~al.(2018)Ying, He, Chen, Eksombatchai, Hamilton, and
  Leskovec]{ying2018graph}
Rex Ying, Ruining He, Kaifeng Chen, Pong Eksombatchai, William~L Hamilton, and
  Jure Leskovec.
\newblock Graph convolutional neural networks for web-scale recommender
  systems.
\newblock In \emph{PKDD}, 2018.

\bibitem[Zafar et~al.(2017{\natexlab{a}})Zafar, Valera, Gomez~Rodriguez, and
  Gummadi]{zafar2017fairness}
Muhammad~Bilal Zafar, Isabel Valera, Manuel Gomez~Rodriguez, and Krishna~P
  Gummadi.
\newblock Fairness beyond disparate treatment \& disparate impact: Learning
  classification without disparate mistreatment.
\newblock In \emph{WWW}, 2017{\natexlab{a}}.

\bibitem[Zafar et~al.(2017{\natexlab{b}})Zafar, Valera, Rodriguez, Gummadi, and
  Weller]{zafar2017parity}
Muhammad~Bilal Zafar, Isabel Valera, Manuel~Gomez Rodriguez, Krishna~P Gummadi,
  and Adrian Weller.
\newblock From parity to preference-based notions of fairness in
  classification.
\newblock \emph{arXiv}, 2017{\natexlab{b}}.

\bibitem[Zhang and Zitnik(2020)]{zhang2020gnnguard}
Xiang Zhang and Marinka Zitnik.
\newblock {GNNguard}: Defending graph neural networks against adversarial
  attacks.
\newblock In \emph{NeurIPS}, 2020.

\bibitem[Zhu et~al.(2019)Zhu, Zhang, Cui, and Zhu]{zhu2019robust}
Dingyuan Zhu, Ziwei Zhang, Peng Cui, and Wenwu Zhu.
\newblock Robust graph convolutional networks against adversarial attacks.
\newblock In \emph{KDD}, 2019.

\bibitem[Zitnik et~al.(2018)Zitnik, Agrawal, and Leskovec]{zitnik2018modeling}
Marinka Zitnik, Monica Agrawal, and Jure Leskovec.
\newblock Modeling polypharmacy side effects with graph convolutional networks.
\newblock In \emph{Bioinformatics}, 2018.

\bibitem[Z{\"u}gner and G{\"u}nnemann(2019)]{zugner2019adversarial}
Daniel Z{\"u}gner and Stephan G{\"u}nnemann.
\newblock Adversarial attacks on graph neural networks via meta learning.
\newblock In \emph{ICLR}, 2019.

\end{thebibliography}


\clearpage
\appendix
\section{Proposition 1 and its Proof}
\label{sec:app_proof_1}

\xhdr{Proposition 1 (Counterfactual Fairness of Downstream Classifier)} \textit{If the representations learned by our framework \nifty satisfy counterfactual fairness, then a downstream classifier $f: \mathbf{z}_{u} \to \hat{y}_{u}$ which leverages these representations also satisfies counterfactual fairness.
}

\textit{Proof.~}
The downstream classifier uses the representation $\mathbf{z}_{u}$ output by our framework for predicting the label $\hat{y}_{u}$ of node $u$, thus forming a Markov chain $\mathbf{x}_{u} \to \mathbf{z}_{u} \to \hat{y}_{u}$ \citep{liao2019learning}.
As we discuss in Section~\ref{sec:prelims}, node representations are said to be counterfactually fair if they are independent of the sensitive attribute. 
\ie the mutual information between the sensitive attribute $s$ and the representation $\mathbf{z}_{u}$ for any given node $u$ is zero: $~I(s; \mathbf{z}_{u})=0$.

Using the properties of inequality and non-negativity of mutual information:
\begin{equation}
    0 \leq I(s; \hat{y}_{u}) \leq I(s; \mathbf{z}_{u}) \text{ and } I(s; \mathbf{z}_{u}){=}0 \implies I(s; \hat{y}_{u}){=}0
\end{equation}
Therefore, the node label $\hat{y}_{u}$ for any given node $u$ is independent of the sensitive attribute $s$, and consequently the downstream node classifier satisfies counterfactual fairness.



\section{Dataset details}
\label{sec:app_dataset}
\xhdr{German Credit Graph}~ The German Graph credit dataset classifies people described by a set of attributes as good or bad credit risks \citep{Dua:2019}.
It consists of attributes like Gender, LoanAmount, and other account-related features of 1,000 clients.
We use Minkowski distance as the similarity measure for calculating the similarity between two node attributes using: $1/(1+\textrm{minkowski}(\mathbf{x}_{u},\mathbf{x}_{v}))$.
To obtain the credit graph network that connects clients, we connect two nodes if the similarity between them is 80\% of the maximum similarity between all respective nodes (Refer Table.~\ref{tab:app_graph} for details).
We argue that a graph neural network is fair if it predicts the client credit risk irrespective of their gender.
Hence, we used \textit{gender} as the sensitive attribute for the loan dataset.

\xhdr{Recidivism Graph} The dataset consists of samples of bail outcomes collected from several state courts in the US between 1990-2009 \citep{jordan2015effect}. 
It consists of past criminal records, demographic attributes, and other
details of 18,876 defendants who got released on bail.
We use Minkowski distance as the similarity measure for calculating the similarity between two node attributes using: $1/(1+\textrm{minkowski}(\mathbf{x}_{u},\mathbf{x}_{v}))$.
To obtain the bail graph network that connects defendants, we connect two nodes if the similarity between them is 60\% of the maximum similarity between all respective nodes (Refer Table.~\ref{tab:app_graph} for details).
A machine learning model is trained to predict a defendant who is more likely to commit a violent or nonviolent crime once released on bail.
A fair model should make predictions independent of the defendant's \textit{race}, and, thus, we use it as the protected attribute for the dataset.

\xhdr{Credit Defaulter Graph} We use a processed version \citep{ustun2019actionable} of the credit dataset in \cite{yeh2009comparisons}.
The task is to predict whether an applicant will default on an upcoming credit card payment.
The dataset contains 30,000 individuals with features like education, credit history, age, and features derived from their spending and payment patterns.
We use Minkowski distance as the similarity measure for calculating the similarity between two node attributes using: $1/(1+\textrm{minkowski}(\mathbf{x}_{u},\mathbf{x}_{v}))$.
To obtain the credit defaulter graph network that connects applicants, we connect two nodes if the similarity between them is 70\% of the maximum similarity between all respective nodes (Refer Table.~\ref{tab:app_graph} for details).
For the credit dataset, we used \textit{age} as the sensitive attribute.


\begin{table*}[h]
\caption{Statistics of novel graph datasets designed for node classification and accompanied by sensitive attributes. The datasets are appropriate to study fairness- and stability-aware algorithms. 
}
\label{tab:app_graph}
\centering
\begin{tabular}{lccc}
\toprule
\multicolumn{1}{c}{\textbf{Dataset}} & \textbf{German credit graph} & \textbf{Recidivism graph} & \textbf{Credit defaulter graph}\\
\toprule
Nodes & 1,000 & 18,876 & 30,000 \\
Edges & 22,242 & 321,308 & 1,436,858\\
Node features & 27 & 18 & 13 \\
Average node degree & 44.48\std{26.51} & 34.04\std{46.65} & 95.79\std{85.88} \\
Sensitive attribute & Gender (Male/Female) & Race (Black/White) & Age ($\leq 25/ >25$)\\
Node labels & good credit vs.~bad credit & bail vs.~no bail & payment default vs.~no default \\
\bottomrule
\end{tabular}
\end{table*}

\section{Architecture and Hyperparameter selection}
\label{sec:app_training}
We provide an overview of the important components of our proposed architecture and their respective training settings.

\xhdr{Encoder}
The encoder block of our proposed framework can comprise of either simple Multilayer Perceptron (MLP) networks or any other GNN variant.
For all our experiments, we use the vanilla GNN as the encoder block of our contrastive learning framework.
For all datasets, we use a single-layer GNN encoder and set the hidden dimensionality to 16.
The encoder is followed by a two-layer MLP projection head \citep{chen2020simple}.
We only use ReLU and BatchNormalization (BN) layers after the first hidden layer in the MLP.
For both the MLP layers, we set the hidden dimensionality to 16.

\xhdr{Predictor} We use a single layer MLP with no ReLU and BN as our predictor \citep{chen2020simple} to transform the graph embeddings of one augmented graph to another and vice-versa.
We set the hidden dimensionality to 16 for the predictor layer.

\xhdr{Downstream classifier} We use a single fully-connected layer with a Sigmoid activation function in all our node-classification experiments. 
We set the hidden dimensionality of the fully-connected layer to 16.

\xhdr{Hyperparameters} For all experiments, we set the probability of perturbing a feature dimension to $p_{n}=0.1$ and the probability with which an edge is dropped to $p_{e}=0.001$.
For training GNNs and their \nifty-augmented counterparts (Sec.~\ref{sec:exp_setup}), we use an Adam optimizer with a learning rate of $1\times10^{-3}$, weight decay of $1\times10^{-5}$, and the number of epochs to $1000$.
For RobustGCN and FairGCN, all hyperparameters are set following the authors’ guidelines.
\begin{figure*}[t]
    \centering
    \begin{flushleft}
	    \small
		\hspace{1.6cm}(a) German Credit Graph
		\hspace{2.7cm}(b) Recidivism Graph
		\hspace{2.4cm}(c) Credit Defaulter Graph
	\end{flushleft}
    \includegraphics[width=0.99\textwidth]{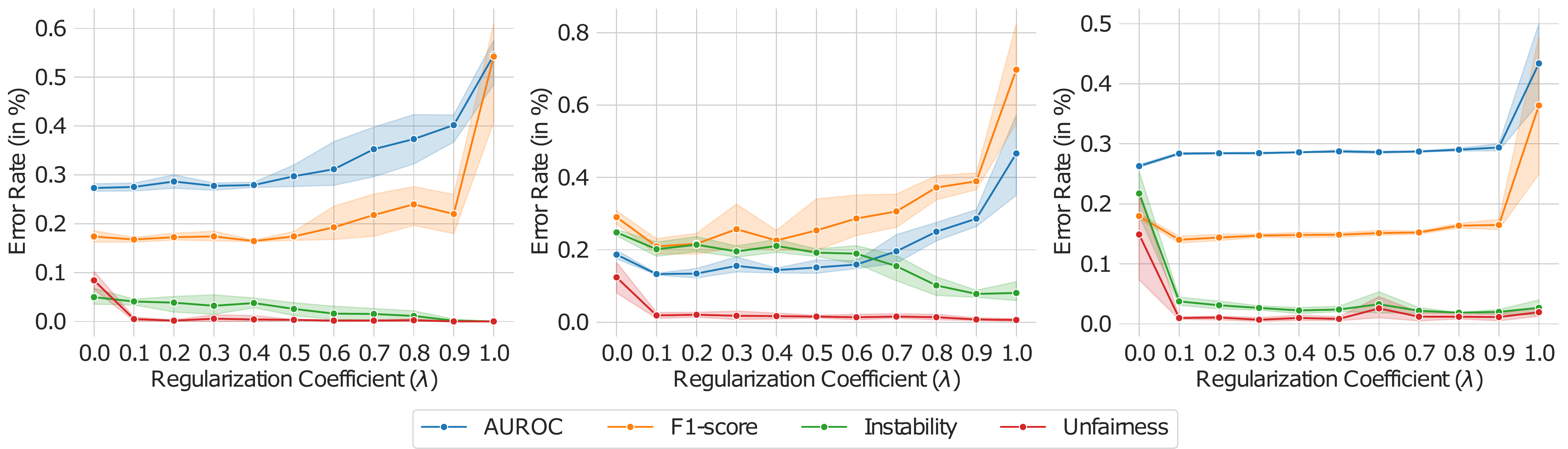}
    \caption{
    Effect of regularization coefficient on AUROC, F1-score, stability, and fairness in \nifty-GIN on (a) the German credit graph, (b) the recidivism graph, and (c) the credit defaulter graph.
    With increasing the regularization coefficient on the self-supervised task the robustness and fairness score can reach $0\%$ error.
    }
    \label{fig:app_tradeoff}
\end{figure*}


\providecommand{\upGamma}{\Gamma}
\providecommand{\uppi}{\pi}

\end{document}